# Follow the Compressed Leader:
# Faster Online Learning of Eigenvectors and Faster MMWU

(version 3)*

Zeyuan Allen-Zhu
zeyuan@csail.mit.edu
Microsoft Research

Yuanzhi Li
yuanzhil@cs.princeton.edu
Princeton University

January 6, 2017

## Abstract

The online problem of computing the top eigenvector is fundamental to machine learning. In both adversarial and stochastic settings, previous results (such as *matrix multiplicative weight update, follow the regularized leader, follow the compressed leader, block power method*) either achieve optimal regret but run slow, or run fast at the expense of loosing a $\sqrt{d}$ factor in total regret where $d$ is the matrix dimension.

We propose a *follow-the-compressed-leader (FTCL)* framework which achieves optimal regret without sacrificing the running time. Our idea is to "compress" the matrix strategy to dimension 3 in the adversarial setting, or dimension 1 in the stochastic setting. These respectively resolve two open questions regarding the design of optimal and efficient algorithms for the online eigenvector problem.

## 1 Introduction

Finding leading eigenvectors of symmetric matrices is one of the most primitive problems in machine learning. In this paper, we study the *online* variant of this problem, which is a learning game between a player and an adversary [2, 17, 20, 26, 30].

**Online (Adversarial) Eigenvector Problem.** The player plays $T$ unit-norm vectors $w_1, \ldots, w_T \in \mathbb{R}^d$ in a row; after playing $w_k$, the adversary picks a feedback matrix $\mathbf{A}_k \in \mathbb{R}^{d \times d}$ that is symmetric and satisfies $0 \preceq \mathbf{A}_k \preceq \mathbf{I}$.[1] Both these assumptions are for the sake of simplicity and can be relaxed.[2] The player then receives a gain

$$w_k^\top \mathbf{A}_k w_k = \mathbf{A}_k \bullet w_k w_k^\top \in [0, 1] \ .$$

The regret minimization problem asks us the player to design a strategy to minimize *regret*, that is, the difference between the total gain obtained by the player and that by the *a posteriori*

---

*The arXiv version appeared on January 6, 2017. This second and third versions polish writing and fix typos.

[1] We denote by $\mathbf{A} \succeq \mathbf{B}$ spectral dominance that is equivalent to saying that $\mathbf{A} - \mathbf{B}$ is positive semidefinite (PSD).

[2] Firstly, all the results cited and stated in this paper, after scaling, generalize to the scenario when the eigenvalues of $\mathbf{A}_k$ are in the range $[l, r]$ for arbitrary $l, r \in \mathbb{R}$. For notational simplicity, we have assumed $l = 0$ and $r = 1$ in this paper. Secondly, if $\mathbf{A}_k$ is not symmetric or even rectangular, classical reductions can turn such a problem into an equivalent online game with only symmetric matrices (see Sec 2.1 of [20]).

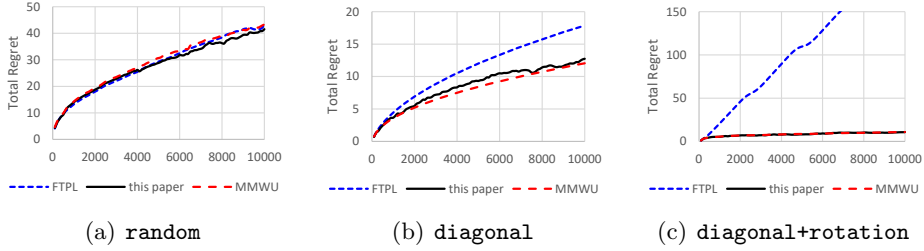

(a) `random`     (b) `diagonal`     (c) `diagonal+rotation`

Figure 1: We generate synthetic data to verify that the total regret of FTPL can indeed be poorer than MMWU or our FTCL. We explain how matrices $\mathbf{A}_k$ are chosen in Appendix A. We have $d = 100$ and the $x$-axis represents the number of iterations.

best fixed strategy $u \in \mathbb{R}^d$:

$$\text{minimize} \quad \max_{u \in \mathbb{R}^d} \sum_{k=1}^T \mathbf{A}_k \bullet (uu^\top - w_k w_k^\top)$$
$$= \lambda_{\max}(\mathbf{A}_1 + \cdots + \mathbf{A}_T) - \sum_{k=1}^T w_k^\top \mathbf{A}_k w_k \ .$$

The name comes from the fact that the player chooses only vectors in a row, but wants to compete against the leading eigenvector in hindsight. To make this problem meaningful, the feedback matrix $\mathbf{A}_k$, is *not* allowed to depend on $w_k$ but can depend on $w_1, \ldots, w_{k-1}$.

## 1.1 Known Results

The most famous solution to the online eigenvector problem is the *matrix multiplicative-weight-update (MMWU)* method, which has also been used towards efficient algorithms for SDP, balanced separators, Ramanujan sparsifiers, and even in the proof of QIP = PSPACE.

**MMWU.** At iteration $k$, define $\mathbf{W}_k = \frac{\exp(\eta \mathbf{\Sigma}_{k-1})}{\mathbf{Tr} \exp(\eta \mathbf{\Sigma}_{k-1})}$ where $\mathbf{\Sigma}_{k-1} \stackrel{\text{def}}{=} \mathbf{A}_1 + \cdots + \mathbf{A}_{k-1}$ and $\eta > 0$ is the learning rate. Then, compute its eigendecomposition

$$\mathbf{W}_k = \frac{\exp(\eta \mathbf{\Sigma}_{k-1})}{\mathbf{Tr} \exp(\eta \mathbf{\Sigma}_{k-1})} = \sum_{j=1}^d p_j \cdot y_j y_j^\top$$

where vectors $y_j$ are normalized eigenvectors. Now, the MMWU strategy instructs the player to choose $w_k = y_j$ each with probability $p_j$. The best choice $\eta = \sqrt{\log d}/\sqrt{T}$ yields a total expected regret $O(\sqrt{T \log d})$ [32], and this is optimal up to constant [11]. It requires some additional, but standard, effort to turn this into a high-confidence result.

Unfortunately, the per-iteration running time of MMWU is at least $O(d^\omega)$ due to eigendecomposition, where $d^\omega$ is the complexity for multiplying two $d \times d$ matrices.[3]

**MMWU-JL.** Some researchers also use the Johnson-Lindenstrauss (JL) compression to reduce the dimension of $\mathbf{W}_k$ from MMWU to make it more efficiently computable [5, 9, 27, 35]. Specifically, they compute a sketch matrix $\mathbf{Y} = \mathbf{W}_k^{1/2} \mathbf{Q}$ using a random $\mathbf{Q} \in \mathbb{R}^{d \times m}$, and then use $\mathbf{Y}\mathbf{Y}^\top$ to approximate $\mathbf{W}_k$. If the dimension $m$ is $\widetilde{O}(1/\sigma^2)$, this compression incurs an average regret loss of $\sigma$. We call this method MMWU-JL for short.[4]

Unfortunately, to maintain a total regret $\widetilde{O}(\sqrt{T})$, one must let $\sigma \approx T^{-1/2}$. Therefore, JL compresses the matrix exponential to dimension $\widetilde{O}(T)$, and is only useful when $T \leq d$.

---

[3]In fact, it is known that eigendecomposition has complexity $O(d^\omega)$ when all the eigenvalues are distinct, and could possibly go up to $O(d^3)$ when some eigenvalues are equal [34].

[4]Through the paper, we use the $\widetilde{O}$ notation to hide polylogarithmic factors in $T, d$ and $1/\varepsilon$ if applicable.

| Paper | Total Regret | Time Per Iteration | Minimum Total Time for $\varepsilon$ Average Regret[5] |
|---|---|---|---|
| MMWU [9, 11] | $\widetilde{O}(\sqrt{T})$ | at least $O(d^\omega)$ | $\widetilde{O}\big(\frac{d^\omega}{\varepsilon^2}\big)$ |
| MMWU-JL [9, 35] ($T \leq d$ only) | $\widetilde{O}(\sqrt{T})$ | $\mathsf{M}^{\mathsf{exp}} \times \widetilde{O}(T)$ | $\widetilde{O}\big(\frac{1}{\varepsilon^{4.5}}\mathsf{nnz}(\mathbf{\Sigma})\big)$ |
| FTPL ($T \geq d$ only) [20] | $\widetilde{O}(\sqrt{dT})$ | $\mathsf{M}^{\mathsf{ev}} \times 1$ | $\widetilde{O}\big(\frac{d^{1.5}}{\varepsilon^{3.5}}\mathsf{nnz}(\mathbf{\Sigma})\big)$ |
| **this paper**    Theorem 1&2 | $\widetilde{O}(\sqrt{T})$ | $\mathsf{M}^{\mathsf{lin}} \times \widetilde{O}(1)$    Theorem 3 | $\widetilde{O}\big(\frac{1}{\varepsilon^{2.5}}\mathsf{nnz}(\mathbf{\Sigma})\big)$ and $\widetilde{O}\big(\frac{1}{\varepsilon^{2.5}}\mathsf{nnz}(\mathbf{\Sigma})^{\frac{3}{4}}\mathsf{nnz}(\mathbf{A})^{\frac{1}{4}} + \frac{1}{\varepsilon^2}\mathsf{nnz}(\mathbf{\Sigma})\big)$ |
| ↓ stochastic online eigenvector only ↓ | | | |
| block power method [20] | $\widetilde{O}(\sqrt{T})$ | $O\big(\mathsf{nnz}(\mathbf{\Sigma})\big)$ | $\widetilde{O}\big(\frac{1}{\varepsilon^2}\mathsf{nnz}(\mathbf{\Sigma})\big)$ |
| **this paper**    Theorem 4 | $\widetilde{O}(\sqrt{T})$ | $O\big(\mathsf{nnz}(\mathbf{A})\big)$    Theorem 4 | $\widetilde{O}\big(\frac{1}{\varepsilon^2}\mathsf{nnz}(\mathbf{A})\big)$ |

Table 1: Comparison of known methods for the online eigenvector problem. We denote by $\mathsf{nnz}(\mathbf{M})$ the time needed to multiply $\mathbf{M}$ to a vector, by $\mathbf{\Sigma} = \mathbf{A}_1 + \cdots + \mathbf{A}_T$, and by $\mathsf{nnz}(\mathbf{A}) = \max_{k \in [T]}\{\mathsf{nnz}(\mathbf{A}_k)\} \leq \mathsf{nnz}(\mathbf{\Sigma})$.
- $\mathsf{M}^{\mathsf{exp}}$ is the time to compute $e^{-\mathbf{M}}$ multiplied with a vector, where $\mathbf{M} \in \mathbb{R}^{d \times d}$ satisfies $0 \preceq \mathbf{M} \preceq \widetilde{O}(T^{1/2}) \cdot \mathbf{I}$.
- $\mathsf{M}^{\mathsf{ev}}$ is the time to compute the leading eigenvector of $\mathbf{M}$ to multiplicative accuracy $O(T^{-3/2}d^{1/2}) \in (0,1)$.
- $\mathsf{M}^{\mathsf{lin}}$ is the time to solve a linear system in $\mathbf{M} \in \mathbb{R}^{d \times d}$, where $\mathbf{M}$ is PSD and of condition number $\leq \widetilde{O}(T^{1/2})$.
- If using iterative methods, the worst-case values $\mathsf{M}^{\mathsf{ev}}, \mathsf{M}^{\mathsf{exp}}, \mathsf{M}^{\mathsf{lin}}$ are (see Section 3)

$$\mathsf{M}^{\mathsf{ev}} = \widetilde{O}\big(\min\{T^{\frac{3}{4}}d^{-\frac{1}{4}}\mathsf{nnz}(\mathbf{\Sigma}), d^\omega\}\big) \geq \mathsf{M}^{\mathsf{exp}} = \widetilde{O}\big(\min\{T^{\frac{1}{4}}\mathsf{nnz}(\mathbf{\Sigma}), d^\omega\}\big) \geq \mathsf{M}^{\mathsf{lin}} = \widetilde{O}\big(\min\{\min\{d, T^{\frac{1}{4}}\}\mathsf{nnz}(\mathbf{\Sigma}), d^\omega\}\big) \ ,$$

where $d^\omega$ is the time needed to multiply two $d \times d$ matrices. If using stochastic iterative methods, $\mathsf{M}^{\mathsf{lin}}$ can be further reduced to $\widetilde{O}\big(T^{\frac{1}{4}}\mathsf{nnz}(\mathbf{\Sigma})^{\frac{3}{4}}\mathsf{nnz}(\mathbf{A})^{\frac{1}{4}} + \mathsf{nnz}(\mathbf{\Sigma})\big)$.

**FTPL.** Researchers also study the *follow-the-perturbed-leader (FTPL)* strategy [2, 17, 20, 26]. In particular, Garber, Hazan and Ma [20] proposed to compute an (approximate) leading eigenvector of the matrix $\mathbf{\Sigma}_{k-1} + rr^\top$ at iteration $k$, where $r$ is a random vector whose norm is around $\sqrt{dT}$.

Unfortunately, the total regret of FTPL is $\widetilde{O}(\sqrt{dT})$, which is a factor $\sqrt{d}$ worse than the optimum regret, and interesting only when $T \geq d$. This factor $\sqrt{d}$ loss can indeed be realized in practice, see Figure 1. In theory, this $d^{1/2}$ factor loss is necessary at least for their proposed method [22].

## 1.2 Our Main Results

We propose a *follow-the-compressed-leader (FTCL)* strategy that, at a high level, compresses the MMWU strategy only to dimension $m = 3$ as opposed to dimension $m = \widetilde{\Theta}(T)$ in MMWU-JL. Our FTCL strategy has significant advantages over previous results because:

- FTCL has regret $\widetilde{O}(\sqrt{T})$ which is optimal up to poly-log factors (as opposed to $\sqrt{d}$ in FTPL).
- Each iteration of FTCL is dominated by solving a logarithmic number of linear systems.

Since solving linear systems is generally no slower than computing eigenvectors or matrix exponentials, the per-iteration complexity of FTCL is no slower than FTPL, and much faster than MMWU and MMWU-JL. We shall make this comparison more explicit in Section 3.

---
[5]The total time complexity of the first $T_\varepsilon$ rounds where $T_\varepsilon$ is the earliest round to achieve an $\varepsilon$ average regret.

## 1.3 Our Side Result: Stochastic Online Eigenvector

We also study the *special case* of the online eigenvector problem where the adversary is *stochastic*, meaning that $\mathbf{A}_1, \ldots, \mathbf{A}_T$ are chosen i.i.d. from a common distribution whose expectation equals $\mathbf{B} \in \mathbb{R}^{d \times d}$, independent of the player's actions. For this problem, there are two goals:

(G1) minimizing regret: $T \cdot \lambda_{\max}(\mathbf{B}) - \sum_{k=1}^{T} w_k^\top \mathbf{A}_k w_k$, and

(G2) finding a unit vector $w$ satisfying $w^\top \mathbf{B} w \geq \lambda_{\max}(\mathbf{B}) - \varepsilon$.

Goal (G1) is obviously stronger than goal (G2). By (martingale) concentration, the quantity $\sum_{k=1}^{T} w_k^\top \mathbf{A}_k w_k$ must be close to $\sum_{k=1}^{T} w_k^\top \mathbf{B} w_k$, and thus given a strategy which minimizes regret for (G1), we can select $w = w_k$ for a uniform random $k \in \{1, \ldots, T\}$, and $w$ should achieve goal (G2) with small $\varepsilon$. In particular, an $\widetilde{O}(\sqrt{T})$ optimal-regret strategy for (G1) gives rise to an algorithm for (G2) with $\varepsilon \leq \widetilde{O}(1/\sqrt{T})$ [4].

Before our work, for this problem:

- Garber *et al.* [20] showed a block power method gives total regret $O(\sqrt{T \log(dT)})$, and runs in $O(\mathsf{nnz}(\boldsymbol{\Sigma}_k))$ time each iteration $k$. (We denote $\mathsf{nnz}(\mathbf{M})$ the time to multiply $\mathbf{M}$ to a vector.)

- Shamir [37] showed Oja's algorithm[6] achieves the weaker goal (G2) with error $\varepsilon = \widetilde{O}(\sqrt{d}/\sqrt{T})$, a factor $\sqrt{d}$ worse than optimum. Oja's algorithm runs in $O(\mathsf{nnz}(\mathbf{A}_k))$ time each iteration $k$.

- It was asked explicitly as an open question in [23] that whether Oja's algorithm can achieve the optimum error $\varepsilon = \widetilde{O}(1/\sqrt{T})$ for goal (G2).[7]

In this paper, we show that

> - Oja's algorithm has total regret $O(\sqrt{T} \log d)$ for the stochastic online eigenvector problem.

This is optimal up to a $\sqrt{\log d}$ factor. It implies an error $\varepsilon = \widetilde{O}(1/\sqrt{T})$ for the easier goal (G2), and thus answers the open question of [23]. Our proof relies on a compression view of Oja's algorithm which compresses MMWU to dimension $m = 1$. Our proof is one-paged, indicating that FTCL might be a better framework for designing and analyzing online algorithms for matrices.

*Remark* 1.1. Our result on Oja's algorithm gives rise to a new framework for non-convex stochastic optimization. One can view each $\mathbf{A}_k$ as a stochastic sample of the negative Hessian matrix, and apply Oja's algorithm to find the most negative eigenvector. This leads to an online algorithm that outperforms stochastic gradient descent with applications to deep learning and so on. [4]

## 1.4 Our Results in a More Refined Language

Denoting by $\lambda \stackrel{\text{def}}{=} \frac{1}{T} \lambda_{\max}(\mathbf{A}_1 + \cdots \mathbf{A}_T)$, we have $\lambda \leq 1$ according to the normalization $\mathbf{A}_k \preceq \mathbf{I}$. In general, the smaller $\lambda$ is, the better a learning algorithm should behave. In the previous subsections, we have followed the tradition and discussed our results and prior works assuming the *worst* possibility of $\lambda$. This has indeed simplified notations.

If $\lambda$ is much smaller than 1, our complexity bounds can be improved to quantities that depend on $\lambda$. We call this the *$\lambda$-refined language*. At a high level, for our FTCL, in both the adversarial and stochastic settings, the total regret improves from $\widetilde{O}(\sqrt{T})$ to $\widetilde{O}(\sqrt{\lambda T})$.

---

[6]Here is a simple description of Oja's algorithm [31]: beginning with a random Gaussian vector $u \in \mathbb{R}^d$, at each iteration $k$, choose $w_k$ to be $(\mathbf{I} + \eta \mathbf{A}_{k-1}) \cdots (\mathbf{I} + \eta \mathbf{A}_1) u$ after normalization.

[7]Jain et al. [23] obtained an optimum formula for $\varepsilon$ when $\mathbf{B}$ has a large gap between its first two leading eigenvalues; and they raised it as an open question in the case without any eigengap assumption. In a separate and earlier work of us [8], we showed that in the special case of $\mathbf{A}_k$ being rank-1, the $\varepsilon = \widetilde{O}(1/\sqrt{T})$ error for Oja's algorithm can indeed by obtained, using very different techniques from this paper.

There is an information-theoretic lower bound of $\Omega(\sqrt{\lambda T})$ for the total regret in this $\lambda$-refined language, see Appendix J. This lower bound even holds for the simpler stochastic online eigenvector problem, even when the matrices $\mathbf{A}_k$ are of rank 1.

As for prior work, it has been recorded that (cf. Theorem 3.1 of [9]) the MMWU and MMWU-JL methods have total regret $O(\sqrt{\lambda T \log d})$. The block power method (for the stochastic setting) has total regret $\widetilde{O}(\sqrt{\lambda T})$, by modifying the proof in [20]. To the best of our knowledge, FTPL has not been analyzed in the $\lambda$-refined language. This may not be a surprise because not all algorithms can take advantage of the $\lambda$-refined language (see for instance one hard instance in bandit problems [3]).

We compare our results with prior work in Table 2 for this $\lambda$-refined language.

## 1.5 Other Related Works

The multiplicative weight update (MWU) method is a simple but powerful algorithmic tool that has been repeatedly discovered in theory of computation, machine learning, optimization, and game theory (see for instance the survey [11] and the book [15]). Its natural matrix extension, matrix-multiplicative-weight-update (MMWU) [32], has been used towards efficient algorithms for solving semidefinite programs [5, 12, 35], balanced separators [33], Ramanujan sparsifiers [9, 27], and even in the proof of QIP = PSPACE [24]. Some authors also refer to MMWU as the *follow-the-regularized-leader (FTRL)* strategy, because MMWU can be analyzed from a mirror-descent view with the matrix entropy function as its regularizer [9].

For the online eigenvector problem, if the feedback matrices $\mathbf{A}_k$ are only of rank-1, the $\widetilde{O}(\sqrt{dT})$ total regret of FTPL can be improved to $\widetilde{O}(d^{1/4}T^{1/2})$. This is first shown by Dwork *et al.* [17] and independently by Kotłowski and Warmuth [26]. However, this $d^{1/4}$ factor for the rank-1 case and the $d^{1/2}$ factor for the high-rank case are tight at least for their proposed FTPL methods [22]. Abernethy *et al.* showed FTPL strategies can be analyzed using a FTRL framework [1].

Researchers also put efforts to understand *high-rank variants* of the online eigenvector problem, that is to obtain the top $k$ eigenvectors as opposed to the top one. Nie *et al.* [30] studied the high-rank adversarial setting simply using MMWU, and their per-iteration complexity is high due to eigendecomposition. Some authors study a very different online model for computing the top $k$ eigenvectors [13, 25]: they wish to output $O(k \cdot \mathsf{poly}(1/\varepsilon))$ vectors instead of $k$ but with a good PCA reconstruction error. In the stochastic setting, the two papers [8, 21] achieve a variant of goal (G2) for finding the top $k$ eigenvectors; however, their techniques are very different from ours and do not imply our results anyways.

For the less relevant *offline* setting, we refer interested readers to our papers [6] (for PCA / SVD) and [7] (for CCA and generalized eigendecomposition) for the most efficient algorithms.

## 1.6 Roadmap

We introduce notations in Section 2, and compare the per-iteration complexity of FTCL to prior work in Section 3. We discuss high-level intuitions and techniques in Section 4. We introduce a new trace inequality in Section 5, and prove our main FTCL result for an oblivious adversary in Section 6. We extend it to the adversarial setting in Section 7, and discuss how to implement FTCL fast in Section 8. Finally, in Section 9 we provide our FTCL result for a stochastic adversary.

Our results are stated directly in the $\lambda$-refined language.

## 2 Notations and Preliminaries

Define $\boldsymbol{\Sigma}_k \stackrel{\text{def}}{=} \sum_{i=1}^k \mathbf{A}_i$ for every $k = 0, 1, \ldots, T$. Since each $\mathbf{A}_k$ is positive semi-definite (PSD), we can find $\mathbf{P}_k \in \mathbb{R}^{d \times d}$ such that $\mathbf{A}_k = \mathbf{P}_k \mathbf{P}_k^\top$; we only use $\mathbf{P}_k$ for analysis purpose only. Given two matrices $\mathbf{A}, \mathbf{B} \in \mathbb{R}^{d \times d}$, we write $\mathbf{A} \bullet \mathbf{B} \stackrel{\text{def}}{=} \mathrm{Tr}(\mathbf{A}^\top \mathbf{B})$. We write $\mathbf{A} \succeq \mathbf{B}$ if $\mathbf{A}, \mathbf{B}$ are symmetric matrices and $\mathbf{A} - \mathbf{B}$ is PSD. We write $[\mathbf{A}]_{i,j}$ the $(i,j)$-th entry of $\mathbf{A}$. We use $\|\mathbf{M}\|_2$ to denote the spectral norm of a matrix $\mathbf{M}$. We use $\mathsf{nnz}(\mathbf{M})$ to denote time needed to multiply matrix $\mathbf{M} \in \mathbb{R}^{d \times d}$ with an arbitrary vector in $\mathbb{R}^d$. In particular, $\mathsf{nnz}(\mathbf{M})$ is at most $d$ plus the number of non-zero elements in $\mathbf{M}$. We denote $\mathsf{nnz}(\mathbf{A}) \stackrel{\text{def}}{=} \max_{k \in [T]} \{\mathsf{nnz}(\mathbf{A}_k)\}$.

Suppose $x_1, \cdots, x_t \in \mathbb{R}$ are drawn i.i.d. from the standard Gaussian $\mathcal{N}(0, 1)$, then $\chi = \sum_{i=1}^t x_i^2$ has a chi-squared distribution of $t$-degree freedom. $\chi^{-1}$ is called inverse-chi-squared distribution of $t$-degree freedom. It is known that $\mathbb{E}[\chi^{-1}] = \frac{1}{t-2}$ for $t \geq 3$.

## 3 Detailed Comparison to Prior Work

We compare the per-iteration complexity of our results more closely to prior work.

In the stochastic setting, Oja's method runs in time $\mathsf{nnz}(\mathbf{A}_k)$ for iteration $k$, and therefore is *undoubtedly faster* than the block power method which runs in time $\mathsf{nnz}(\boldsymbol{\Sigma}_k)$.

In the adversarial setting, it is *clear* that the per-iteration complexities of FTPL and FTCL are no greater than MMWU, because computing the leading eigenvector and the matrix inversion are both faster than computing the full eigendecomposition. In the rest of this section, we compare MMWU-JL, FTPL and FTCL more closely. They respectively have per-iteration complexities

$$\widetilde{O}(T) \times \mathsf{M}^{\mathsf{exp}}, \quad 1 \times \mathsf{M}^{\mathsf{ev}}, \quad \text{and} \quad \widetilde{O}(1) \times \mathsf{M}^{\mathsf{lin}}$$

where

- In MMWU-JL, we denote by $\mathsf{M}^{\mathsf{exp}}$ the time needed for computing $\exp(\eta \boldsymbol{\Sigma}_{k-1}/2)$ multiplied to a vector. Recall that $\eta = \widetilde{\Theta}(T^{-1/2})$.
- In FTPL, following the tradition, we denote by $\mathsf{M}^{\mathsf{ev}}$ the time needed for computing the top eigenvector of $\boldsymbol{\Sigma}_{k-1} + rr^\top$, where the norm of $r$ is $O(\sqrt{dT})$.
- In FTCL, we denote by $\mathsf{M}^{\mathsf{lin}}$ the time needed for solving a linear system with matrix $\mathbf{M} = c\mathbf{I} - \eta \boldsymbol{\Sigma}_{k-1}$, where $\mathbf{M} \succeq \frac{1}{e}\mathbf{I}$ and $\eta = \widetilde{\Theta}(T^{-1/2})$.

For exact computations, one may generally derive that $\mathsf{M}^{\mathsf{exp}} \geq \mathsf{M}^{\mathsf{ev}} \geq \mathsf{M}^{\mathsf{lin}}$. However, for large-scale applications, one usually applies iterative methods for the three tasks. Iterative methods utilize matrix sparsity, and have running times that depend on matrix properties.

**Worst-case Complexity.** We compute that:

- $\mathsf{M}^{\mathsf{exp}}$ in the worst case is $\widetilde{O}(\min\{T^{1/4}\mathsf{nnz}(\boldsymbol{\Sigma}_T), d^\omega\})$.

  The first is because if using Chebyshev approximation, one can compute $\exp(\eta \boldsymbol{\Sigma}_{k-1}/2)$ applied to a vector in time at most $\widetilde{O}(\|\eta \boldsymbol{\Sigma}_{k-1}\|_2^{1/2} \cdot \mathsf{nnz}(\boldsymbol{\Sigma}_{k-1}))$. The second is because one can compute the singular value decomposition of $\boldsymbol{\Sigma}_{k-1}$ in time $\widetilde{O}(d^\omega)$ and then compute the matrix $\exp(\eta \boldsymbol{\Sigma}_{k-1}/2)$ directly.

- $\mathsf{M}^{\mathsf{ev}}$ in the worst case is $\widetilde{O}(\min\{T^{3/4} d^{-1/4} \mathsf{nnz}(\boldsymbol{\Sigma}_T), d^\omega\})$.

  The first is so because, as proved in [20], it suffices to compute the top eigenvector of $\boldsymbol{\Sigma}_{k-1} + rr^\top$ up to a multiplicative error $O(T^{-\frac{3}{2}} d^{\frac{1}{2}})$.[8] If one applies Lanczos method, this is in time $\widetilde{O}(T^{\frac{3}{4}}d^{-\frac{1}{4}}\mathsf{nnz}(\mathbf{\Sigma}_T))$. (Recall that it only works when $T \geq d$). The second is because the leading eigenvector of a $d \times d$ matrix can be computed directly in time $O(d^\omega)$.

---

[8] A multiplicative error $\delta$ means to find $x$ such that $x^\top (\boldsymbol{\Sigma}_{k-1} + rr^\top) x \geq (1-\delta) \lambda_{\max}(\boldsymbol{\Sigma}_{k-1} + rr^\top)$.

- $\mathsf{M}^{\mathsf{lin}}$ in the worst case is $\widetilde{O}\big(\min\{\min\{T^{\frac{1}{4}}, d\}\mathsf{nnz}(\mathbf{\Sigma}_T), d^\omega\}\big)$.

  The first is because our matrix $\mathbf{M}$ has a condition number (i.e., $\lambda_{\max}(\mathbf{M})/\lambda_{\min}(\mathbf{M})$) at most $O(\eta T) = \widetilde{O}(T^{1/2})$. If using conjugate gradient [38], one can solve a linear system for $\mathbf{M}$ in time at most $\widetilde{O}\big(\min\{T^{\frac{1}{4}}, d\}\mathsf{nnz}(\mathbf{\Sigma}_T)\big)$. The second is because the inverse of a $d \times d$ matrix can be computed directly in time $O(d^\omega)$ [14].

- $\mathsf{M}^{\mathsf{lin}}$ can be improved to $\widetilde{O}\big(\min\big\{T^{\frac{1}{4}}\mathsf{nnz}(\mathbf{\Sigma}_T)^{\frac{3}{4}}\mathsf{nnz}(\mathbf{A})^{\frac{1}{4}} + \mathsf{nnz}(\mathbf{\Sigma}_T), d^\omega\big\}\big)$ if using stochastic iterative methods.

In sum, if using iterative methods, the worst case values of $\mathsf{M}^{\mathsf{lin}}$, $\mathsf{M}^{\mathsf{ev}}$, $\mathsf{M}^{\mathsf{exp}}$ are on the same magnitude. Since the per-iteration cost of FTCL is only $\widetilde{O}(\mathsf{M}^{\mathsf{lin}})$, this is no slower than $O(\mathsf{M}^{\mathsf{ev}})$ of FTPL, and much faster than $O(T \times \mathsf{M}^{\mathsf{exp}})$ of MMWU-JL.

**Practical Complexity.** There are many algorithms to compute leading eigenvectors, including Lanczos method, shift-and-invert, and the (slower) power method. The performance may depend on other properties of the matrix, including "how well-clustered the eigenvalues are."

There are also numerous ways to compute matrix inversions, including conjugate gradient, accelerated coordinate descent, Chebyshev method, accelerated SVRG, and many others. Some of them also run faster when the eigenvalues form clusters [38].

In particular, for a random Gaussian matrix $\mathbf{\Sigma}_{k-1}$ (with dimension $100 \sim 5000$), using the default scientific package SciPy of Python, $\mathsf{M}^{\mathsf{ev}}$ is roughly 3 times of $\mathsf{M}^{\mathsf{lin}}$.

**Total Worst-Case Complexity.** Since FTPL requires $d$ times *more iterations* in order to achieve the same average regret as FTCL or MMWU, in the last column of Table 1, we also summarize the minimum *total* time complexity needed to achieve an $\varepsilon$ average regret.

> **Examples.** If $\mathsf{nnz}(\mathbf{\Sigma}_T) = d^2$ and $\mathsf{nnz}(\mathbf{A}) = O(d)$, the total complexity needed to achieve an $\varepsilon$ average regret:
> 
> $\widetilde{O}(d^2\varepsilon^{-2} + d^{1.75}\varepsilon^{-2.5})$ (by us)  vs.  $\widetilde{O}(d^2\varepsilon^{-4.5})$ (by MMWU-JL)  or  $\widetilde{O}(d^\omega\varepsilon^{-2})$ (by MMWU) .

**$\lambda$-refined setting.** In the $\lambda$-refined setting, one can revise the complexity bounds accordingly.

For all the three methods FTCL, MMWU and MMWU-JL, the optimal learning rate $\eta$ becomes $\widetilde{O}((\lambda T)^{-1/2})$ in this setting, and they achieve an average $\varepsilon$ regret in at most $T = \widetilde{O}(\lambda/\varepsilon^2)$ iterations. The running time of MMWU therefore improves by a factor of $\lambda$.

As for MMWU-JL, the worst-case value $\mathsf{M}^{\mathsf{exp}}$ is $\widetilde{O}(\|\eta\mathbf{\Sigma}_T\|_2^{1/2} \cdot \mathsf{nnz}(\mathbf{\Sigma}))$ if using conjugate gradient, and this spectral norm $\|\eta\mathbf{\Sigma}_T\|_2 \leq O(\eta\|\mathbf{\Sigma}_T\|) \leq O(\eta\lambda T) = O(\lambda/\varepsilon)$. Moreover, the compressed dimension of MMWU-JL must be $\widetilde{O}(\varepsilon^{-2})$ in order to achieve an $\varepsilon$ average regret. This gives a per-iteration worst-case complexity $\widetilde{O}(\lambda^{1/2}\varepsilon^{-5/2}\mathsf{nnz}(\mathbf{\Sigma}))$ and thus a total complexity of $\widetilde{O}(\lambda^{1.5}\varepsilon^{-4.5}\mathsf{nnz}(\mathbf{\Sigma}))$.

As for our FTCL, the worst-case value $\mathsf{M}^{\mathsf{lin}}$ depends on the condition number of the matrix $\mathbf{M} = c\mathbf{I} - \eta\mathbf{\Sigma}_{k-1}$ we invert at each iteration. The condition number of $\mathbf{M}$ is at most $\|\eta\mathbf{\Sigma}_T\|_2 \leq O(\lambda/\varepsilon)$, so the per-iteration worst-case complexity is $\widetilde{O}(\lambda^{1/2}\varepsilon^{-1/2}\mathsf{nnz}(\mathbf{\Sigma}))$ if using conjugate gradient, and the total complexity is $\widetilde{O}(\lambda^{1.5}\varepsilon^{-2.5}\mathsf{nnz}(\mathbf{\Sigma}))$. Alternatively, if one uses the accelerated SVRG method to compute this inversion, the per-iteration worst-case complexity is $\widetilde{O}\big(\sqrt{\eta T}\mathsf{nnz}(\mathbf{\Sigma})^{\frac{3}{4}}\mathsf{nnz}(\mathbf{A})^{\frac{1}{4}} + \mathsf{nnz}(\mathbf{\Sigma})\big) = \widetilde{O}\big(\varepsilon^{-0.5}\mathsf{nnz}(\mathbf{\Sigma})^{\frac{3}{4}}\mathsf{nnz}(\mathbf{A})^{\frac{1}{4}} + \mathsf{nnz}(\mathbf{\Sigma})\big)$.

## 4 High-Level Discussion of Our Techniques

**Revisit MMWU.** We first revisit the high-level idea behind the proof of MMWU. Recall $\mathbf{W}_k = \exp(c_k\mathbf{I} + \eta\mathbf{\Sigma}_{k-1})$ where $c_k$ is the unique constant such that $\mathbf{Tr}\mathbf{W}_k = 1$. The main proof step (see for instance [9, Theorem 3.1]) is to use the equality $\mathbf{Tr}\mathbf{W}_k = \mathbf{Tr}\mathbf{W}_{k+1} = 1$ to derive a relationship between $c_k - c_{k+1}$ and the gain value $\mathbf{W}_k \bullet \mathbf{A}_k$ at this iteration.

More specifically, using the Golden-Thompson inequality we have

$$\mathbf{Tr}\big(e^{c_k\mathbf{I}+\eta\mathbf{\Sigma}_k}\big) \leq \mathbf{Tr}\big(e^{c_k\mathbf{I}+\eta\mathbf{\Sigma}_{k-1}}e^{\eta\mathbf{A}_k}\big) = \mathbf{Tr}\big(\mathbf{W}_k e^{\eta\mathbf{A}_k}\big) \approx \mathbf{Tr}\big(e^{c_k\mathbf{I}+\eta\mathbf{\Sigma}_{k-1}}\big) + \eta\mathbf{W}_k \bullet \mathbf{A}_k \enspace.$$

One can also use convexity to show

$$\mathbf{Tr}\big(e^{c_{k+1}\mathbf{I}+\eta\mathbf{\Sigma}_k}\big) - \mathbf{Tr}\big(e^{c_k\mathbf{I}+\eta\mathbf{\Sigma}_k}\big) \leq c_{k+1} - c_k \enspace.$$

Adding these two inequalities, and using the fact that $\mathbf{Tr}\mathbf{W}_k = \mathbf{Tr}\mathbf{W}_{k+1} = 1$, we immediately have $c_k - c_{k+1} \lesssim \eta\mathbf{W}_k \bullet \mathbf{A}_k$. In other words, the gain value $\mathbf{W}_k \bullet \mathbf{A}_k$ at iteration $k$, up to a factor $\eta$, is lower bounded by the decrement of $c_k$. On the other hand, it is easy to see $c_1 - c_{T+1} \geq \eta\lambda_{\max}(\mathbf{\Sigma}_T) - O(\log d)$ from $c_1 = -\log d$ and the definition of $c_{T+1}$. Together, we can derive that

$$\sum_{k=1}^T \mathbf{W}_k \bullet \mathbf{A}_k \gtrsim \lambda_{\max}(\mathbf{\Sigma}_T) \enspace.$$

In the rest of this section, we perform a thought experiment to "modify" the above MMWU analysis step-by-step. In the end, the intuition of our FTCL shall become clear to the reader.

**Thinking Step 1.** We wish to choose a random Gaussian vector $u \in \mathbb{R}^d$ and "compress" MMWU to dimension 1 in the direction of $u$. More specifically, we define $\mathbf{W}_k = \exp(c_k\mathbf{I} + \eta\mathbf{\Sigma}_{k-1})$ but this time $c_k$ is the unique constant such that $\mathbf{Tr}(\mathbf{W}_k uu^\top) = u^\top \mathbf{W}_k u = 1$. In such a case, we *wish* to say that

$$\mathbf{Tr}\big(e^{c_k\mathbf{I}+\eta\mathbf{\Sigma}_k}uu^\top\big) = \mathbf{Tr}\big(e^{c_k\mathbf{I}+\eta\mathbf{\Sigma}_{k-1}+\eta\mathbf{A}_k}uu^\top\big) \stackrel{(\star)}{\leq} \mathbf{Tr}\big(e^{(c_k\mathbf{I}+\eta\mathbf{\Sigma}_{k-1})/2}uu^\top e^{(c_k\mathbf{I}+\eta\mathbf{\Sigma}_{k-1})/2}e^{\eta\mathbf{A}_k}\big)$$
$$= \mathbf{Tr}\big(\mathbf{W}_k^{1/2}uu^\top \mathbf{W}_k^{1/2}e^{\eta\mathbf{A}_k}\big) \approx \mathbf{Tr}\big(\mathbf{W}_k uu^\top\big) + \eta\mathbf{W}_k^{1/2}uu^\top\mathbf{W}_k^{1/2} \bullet \mathbf{A}_k \enspace.$$

If the above inequality were true, then we could define $w_k \stackrel{\text{def}}{=} \mathbf{W}_k^{1/2}u$ which is a unit vector (because $\mathbf{Tr}(\mathbf{W}_k uu^\top) = 1$) and the gain $w_k^\top \mathbf{A}_k w_k = w_k w_k^\top \bullet \mathbf{A}_k$ would again be proportional to the change of this new potential function $\mathbf{Tr}\big(e^{c_k\mathbf{I}+\eta\mathbf{\Sigma}_{k-1}}uu^\top\big)$. This idea almost worked except that inequality $(\star)$ is false due to the non-commutativity of matrices.[9]

Perhaps the most "immediate" idea to fix this issue is to use the randomness of $uu^\top$. Recall that $\mathbb{E}[uu^\top] = \mathbf{I}$ if we choose properly normalize $u$, and therefore it "seems like" we have $\mathbb{E}[\mathbf{Tr}(\mathbf{W}_k uu^\top)] = \mathbf{Tr}(\mathbf{W}_k)$ and the inequality will go through. Unfortunately, this idea fails for a fundamental reason: the normalization constant $c_k$ depends on $u$, so $\mathbf{W}_k$ is *not* independent from the randomness of $u$.[10]

**Thinking Step 2.** Since Gaussian vectors are rotationally invariant, we switch wlog to the eigenbasis of $\mathbf{\Sigma}_{k-1}$ so $\mathbf{W}_k$ is a diagonal matrix. We make an important observation:[11]

> $c_k$ depends only on $|u_1|, \ldots, |u_d|$, but not on the $2^d$ possible signs of $u_1, \ldots, u_d$.

For this reason, we can fix a diagonal matrix $\mathbf{D}$ and consider all random $uu^\top$ which *agree* with $\mathbf{D}$

---

[9] A analogy for this effect can be found in the inequality $\mathbf{Tr}(e^\mathbf{A}) \leq \mathbf{Tr}(e^\mathbf{B})$ for every $\mathbf{A} \preceq \mathbf{B}$. This inequality becomes false when multiplied with $uu^\top$ and in general $e^\mathbf{A} \preceq e^\mathbf{B}$ is false.

[10] In fact, $c_k$ can be made almost independent from $u$ if we replace $uu^\top$ with $\mathbf{Q}\mathbf{Q}^\top$ where $\mathbf{Q}$ is a random $d \times m$ matrix for some very large $m$. That was the main idea behind MMWU-JL.

[11] This is because, $\mathbf{Tr}(e^{c_k\mathbf{I}-\eta\mathbf{\Sigma}_{k-1}}uu^\top) = \sum_{i=1}^d \big(|u_i|^2 \cdot e^{c_k - \eta\lambda_i}\big)$ where $\lambda_i$ is the $i$-th eigenvalue of $\mathbf{\Sigma}_{k-1}$.

on its diagonal,[12] All of such vectors $u$ give the same normalization constant $c_k$, and it satisfies $\mathbb{E}[uu^\top|\mathbf{D}] = \mathbf{D}$. This implies that we can now study the conditional expected potential change

$$\mathbb{E}\left[\mathbf{Tr}\left(e^{c_k\mathbf{I}+\eta\boldsymbol{\Sigma}_k}uu^\top\right) - \mathbf{Tr}\left(e^{c_k\mathbf{I}+\eta\boldsymbol{\Sigma}_{k-1}}uu^\top\right)\middle|\mathbf{D}\right] = \mathbf{Tr}\left(e^{c_k\mathbf{I}+\eta\boldsymbol{\Sigma}_k}\mathbf{D}\right) - \mathbf{Tr}\left(e^{c_k\mathbf{I}+\eta\boldsymbol{\Sigma}_{k-1}}\mathbf{D}\right) ,$$

or if we denote by $\mathbf{B} = c_k\mathbf{I} + \eta\boldsymbol{\Sigma}_{k-1}$, we want to study the difference $\mathbf{Tr}(e^{\mathbf{B}+\eta\mathbf{A}_k}\mathbf{D}) - \mathbf{Tr}(e^{\mathbf{B}}\mathbf{D})$ only in the special case that $\mathbf{D}$ and $\mathbf{B}$ are *simultaneously diagonalizable*.

**Thinking Step 3.** A usual way to bound $\mathbf{Tr}(e^{\mathbf{B}+\eta\mathbf{A}_k}\mathbf{D}) - \mathbf{Tr}(e^{\mathbf{B}}\mathbf{D})$ is to define $f(\eta) \stackrel{\text{def}}{=} \mathbf{Tr}(e^{\mathbf{B}+\eta\mathbf{A}_k}\mathbf{D})$ and bound $f(\eta)$ by its Taylor series $f(0) + f'(0)\eta + \frac{1}{2}f''(0)\eta^2 + \cdots$. The zero-order derivative $f(0)$ is $\mathbf{Tr}(e^{\mathbf{B}}\mathbf{D})$. The first-order derivative $f'(0) = \mathbf{Tr}(\mathbf{A}_k e^{\mathbf{B}}\mathbf{D}) = e^{\mathbf{B}/2}\mathbf{D}e^{\mathbf{B}/2} \bullet \mathbf{A}_k$ behaves exactly in the way we hope, and this strongly relies on the commutativity between $\mathbf{B}$ and $\mathbf{D}$. Unfortunately, higher-order derivatives $f^{(k)}(0)$ benefit less and less from the commutativity between $\mathbf{B}$ and $\mathbf{D}$ due to the existence of terms such as $\mathbf{A}_k e^{\mathbf{B}}\mathbf{D}e^{\mathbf{B}}\mathbf{A}_k\mathbf{D}$. For this reason, we need to (1) truncate the Taylor series and (2) use different analytic tools. This motivates us to use the following regime that can be viewed as a "low-degree" version of MMWU:

**A Quick Detour.** In a recent result, the authors of [9] generalized MMWU to $\ell_{1-1/q}$ regularized strategies. For every $q \geq 2$, they define $\mathbf{X}_k = (c_k\mathbf{I} - \eta\boldsymbol{\Sigma}_{k-1})^{-q}$ where $c_k$ is the unique constant such that $c_k\mathbf{I} - \eta\boldsymbol{\Sigma}_{k-1} \succ 0$ and $\mathbf{Tr}\mathbf{X}_k = 1$.[13] This is a generalization of MMWU because when $q \approx \log d$, the matrix $\mathbf{X}_k$ behaves nearly the same as $\mathbf{W}_k$; in particular, it gives the same regret bound. The analysis behind this new strategy is to keep track of the potential change in $\mathbf{Tr}\left((c_k\mathbf{I}-\eta\boldsymbol{\Sigma}_{k-1})^{-(q-1)}\right)$ as opposed to $\mathbf{Tr}\left(e^{c_k\mathbf{I}+\eta\boldsymbol{\Sigma}_{k-1}}\right)$, and then use the so-called Lieb-Thirring inequality (see Section 5) to replace the use of Golden-Thompson. (Note that $c_k$ is choosen with respect to $q$ but the potential is with respect to $q-1$.)

**Thinking Step 4.** Let us now replace MMWU strategies in our Thinking Steps 1,2,3 with $\ell_{1-1/q}$ regularized strategies. Such strategies have two advantages: (1) they help us overcome the issue for higher-order terms in Thinking Step 3, and (2) solving linear systems is *more efficient* than computing matrices exponentials. We shall choose $q = \Theta(\log(dT))$ in the end.

Specifically, we prepare a random vector $u$ and define the normalization constant $c_k$ to be the unique one satisfying $\mathbf{Tr}\left((c_k\mathbf{I} - \eta\boldsymbol{\Sigma}_{k-1})^{-q}uu^\top\right) = \mathbf{Tr}(\mathbf{X}_k uu^\top) = 1$. At iteration $k$, we let the player choose strategy $\mathbf{X}_k^{1/2}u$ which is a unit vector.

If one goes through all the math carefully (using Woodbury formula), this time we are entitled to upper bound the trace difference of the form $\mathbf{Tr}((\mathbf{B}+\eta\mathbf{C})^{q-1}\mathbf{D}) - \mathbf{Tr}(\mathbf{B}^{q-1}\mathbf{D})$ where $\mathbf{D}$ is simultaneously diagonalizable with $\mathbf{B}$ but not $\mathbf{C}$. Similar to Thinking Step 3, we can define $f(\eta) \stackrel{\text{def}}{=} \mathbf{Tr}((\mathbf{B}+\eta\mathbf{C})^{q-1}\mathbf{D})$ and bound this polynomial $f(\eta)$ using its Taylor expansion at point 0. Commutativity between $\mathbf{B}$ and $\mathbf{D}$ helps us compute $f'(0) = (q-1)\mathbf{Tr}(\mathbf{B}^{q-2}\mathbf{C}\mathbf{D})$ but again we cannot bound higher-derivatives directly. Fortunately, this time $f(\eta)$ is a degree $q-1$ polynomial so we can use Markov brothers' inequality to give an upper bound on its higher-order terms. This is the place we lose a few extra polylogarithmic factors in the total regret.

**Thinking Step 5.** Somehow necessarily, even the second-order derivative $f''(0)$ can depend on terms such as $1/D_{ii}$ where $D_{ii} = |u_i|^2$ is the $i$-th diagonal entry of $\mathbf{D}$. This quantity, over the Gaussian random choice of $u_i$, does not have a bounded mean. More generally, the inverse chi-squared distribution with degree $t$ (recall Section 2) has a bounded mean only when $t \geq 3$. For this reason, instead of picking a single random vector $u \in \mathbb{R}^d$, we need to pick three random vectors $u_1, u_2, u_3 \in \mathbb{R}^d$ and replace all the occurrences of $uu^\top$ with $\frac{1}{3}(u_1 u_1^\top + u_2 u_2^\top + u_3 u_3^\top)$ in the previous thinking steps. As a result, each $D_{ii}$ becomes a chi-squared distribution of degree 3 so the issue goes away. This is why we claimed in the introduction that

$$\boxed{\textit{we can compress MMWU to dimension 3.}}$$

---

[12]That is, all random $uu^\top$ such that $\|u_i\|_2^2 = \mathbf{D}_{i,i}$ for each $i \in [d]$. For simplicity we also denote this event as $\mathbf{D}$.

[13]The name "$\ell_{1-1/q}$ strategies" comes from the following fact. Recall MMWU naturally arises as the follow-the-regularized-leader strategy, where the regularizer is the matrix entropy. If the entropy function is replaced with a negative $\ell_{1-1/q}$ norm, the resulting strategy becomes $\mathbf{X}_k$. We encourage interested readers to see the introduction of [9] for more background, but we shall make this present paper self-contained.

REMARK. By losing a polylog factor in regret, one can compress it further to dimension 2. This is because the mean of the inverse chi-squared distribution with degree 2, if truncated at some large value $v$, is only $\log(v)$. However, this "truncated mean" becomes $\Omega(\sqrt{v})$ for degree 1.

**Thinking Step 6.** Putting together previous steps, we obtain a FTCL strategy with total regret $O(\sqrt{T} \log^3(dT))$, which is worse than MMWU only by a factor $O(\log^{2.5}(dT))$. We call this method FTCL$^{\mathsf{obl}}$ and include its analysis in Section 6. However, FTCL$^{\mathsf{obl}}$ only works for an oblivious adversary (i.e., when $\mathbf{A}_1, \ldots, \mathbf{A}_T$ are fixed a priori) and gives an expected regret. To turn it into a robust strategy against *adversarial* $\mathbf{A}_1, \ldots, \mathbf{A}_T$, and to make the regret bound work with high confidence, we need to re-sample $u_1, u_2, u_3$ every iteration. We call this method FTCL$^{\mathsf{adv}}$. A careful but standard analysis with Azuma inequality helps us reduce FTCL$^{\mathsf{adv}}$ to FTCL$^{\mathsf{obl}}$. We state this result in Section 7.

**Running Time.** As long as $q$ is an even integer, the computation of "$(c_k \mathbf{I} - \eta \mathbf{\Sigma}_{k-1})^{-1}$ applied to a vector" (or in other words, solving linear systems) becomes the bottleneck for each iteration of FTCL$^{\mathsf{obl}}$ and FTCL$^{\mathsf{adv}}$. However, as long as $q \geq \Omega(\log(dT))$, we show that the condition number of the matrix $c_k \mathbf{I} - \eta \mathbf{\Sigma}_{k-1}$ is at most $\eta T = \Theta(T^{1/2})$. Conjugate gradient solves each such linear system in worst-case time $\widetilde{O}(\min\{T^{1/4}, d\} \times \mathsf{nnz}(\mathbf{\Sigma}_{k-1}))$.

**Compress to 1-d in Stochastic Online Eigenvector.** If the adversary is stochastic, we observe that Oja's algorithm corresponds to a potential function $\mathbf{Tr}\big((\mathbf{I} + \eta \mathbf{A}_k) \cdots (\mathbf{I} + \eta \mathbf{A}_1) uu^\top (\mathbf{I} + \eta \mathbf{A}_1) \cdots (\mathbf{I} + \eta \mathbf{A}_k)\big)$. Because the matrices are drawn from a common distribution, this potential behaves similar to the matrix exponential but compressed to dimension 1, namely $\mathbf{Tr}\big(e^{\eta(\mathbf{A}_1 + \cdots + \mathbf{A}_k)} uu^\top\big)$. In fact, just using linearity of expectation carefully, one can both upper and lower bound this potential. We state this result in Section 9 (and it can be proved in one page!)

## 5 A New Trace Inequality

Prior work on MMWU and its extensions rely heavily on one of the following trace inequalities [9]:

Golden-Thompson inequality : $\mathbf{Tr}(e^{\mathbf{A} + \eta \mathbf{B}}) \leq \mathbf{Tr}(e^{\mathbf{A}} e^{\eta \mathbf{B}})$

Lieb-Thirring inequality : $\mathbf{Tr}\big((\mathbf{A} + \eta \mathbf{B})^k\big) \leq \mathbf{Tr}\big(\mathbf{A}^{k/2} (\mathbf{I} + \eta \mathbf{A}^{-1/2} \mathbf{B} \mathbf{A}^{-1/2})^k \mathbf{A}^{k/2}\big)$ .

Due to our compression framework in this paper, we need inequalities of type

"$\mathbf{Tr}(e^{\mathbf{A} + \eta \mathbf{B}} \mathbf{D}) \leq \mathbf{Tr}\big(e^{\eta \mathbf{B}} e^{\mathbf{A}/2} \mathbf{D} e^{\mathbf{A}/2}\big)$"

"$\mathbf{Tr}\big((\mathbf{A} + \eta \mathbf{B})^k \mathbf{D}\big) \leq \mathbf{Tr}\big((\mathbf{I} + \eta \mathbf{A}^{-1/2} \mathbf{B} \mathbf{A}^{-1/2})^k \mathbf{A}^{k/2} \mathbf{D} \mathbf{A}^{k/2}\big)$ . "  (5.1)

which look almost like "generalizations" of Golden-Thompson and Lieb-Thirring (by setting $\mathbf{D} = \mathbf{I}$). Unfortunately, such generalizations ***do not hold*** for an arbitrary $\mathbf{D}$. For instance, if the first "generalization" holds for every PSD matrix $\mathbf{D}$ then it would imply " $e^{\mathbf{A} + \eta \mathbf{B}} \preceq e^{\mathbf{A}/2} e^{\eta \mathbf{B}} e^{\mathbf{A}/2}$ " which is a false inequality due to matrix non-commutativity.

In this paper, we show that if $\mathbf{D}$ is *commutative* with $\mathbf{A}$, then the "generalization" (5.1) holds *for the zeroth and first order terms* with respect to $\eta$. As for higher order terms, we can control it using Markov brothers' inequality. (Proof in Appendix B.)

**Lemma 5.1.** *For every symmetric matrices* $\mathbf{A}, \mathbf{B}, \mathbf{D} \in \mathbb{R}^{d \times d}$, *every integer* $k \geq 1$, *every* $\eta^* \geq 0$, *and every* $\eta \in [0, \eta^*/k^2]$, *if* $\mathbf{A}$ *and* $\mathbf{D}$ *are* commutative, *then*

$$(\mathbf{A} + \eta \mathbf{B})^k \bullet \mathbf{D} - \mathbf{A}^k \bullet \mathbf{D} \leq k\eta \mathbf{B} \bullet \mathbf{A}^{k-1} \mathbf{D} + \left(\frac{\eta k^2}{\eta^*}\right)^2 \max_{\eta' \in [0, \eta^*]} \left\{ \left| (\mathbf{A} + \eta' \mathbf{B})^k \bullet \mathbf{D} - \mathbf{A}^k \bullet \mathbf{D} \right| \right\} .$$

# 6 Oblivious Online Eigenvector + Expected Regret

In this section we first focus on a simpler oblivious setting. $\mathbf{A}_1, \ldots, \mathbf{A}_T$ are $T$ PSD matrices chosen by the adversary *in advance*, and they do not depend on the player's actions in the $T$ iterations. We are interested in upper bounding the total *expected* regret

$$\lambda_{\max}\bigl(\textstyle\sum_{k=1}^T \mathbf{A}_k\bigr) - \sum_{k=1}^T \mathbb{E}[w_k^\top \mathbf{A}_k w_k] ,$$

where the expectation is over player's random choices $w_k \in \mathbb{R}^d$ (recall $\|w_k\|_2 = 1$).

In Section 7 we generalize this result to the full adversarial setting along with high-confidence regret.

Our algorithm FTCL$^{\mathsf{obl}}$ is presented in Algorithm 1. It is parameterized by an even integer $q \geq 2$ and a learning rate $\eta > 0$. It initializes with a rank-3 Wishart random matrix $\mathbf{U}$. For every $k \in [T+1]$, we denote by $\mathbf{X}_k \stackrel{\text{def}}{=} (c_k \mathbf{I} - \eta \mathbf{\Sigma}_{k-1})^{-q}$ where[14]

$$c_k > 0 \text{ is the unique constant s.t.} \quad c_k \mathbf{I} - \eta \mathbf{\Sigma}_{k-1} \succ 0 \quad \text{and} \quad \mathbf{Tr}(\mathbf{X}_k \mathbf{U}) = 1 .$$

At iteration $k \in [T]$, the player plays a random unit vector $w_k$, among the three eigenvectors of $\mathbf{X}_k^{1/2} \mathbf{U} \mathbf{X}_k^{1/2}$. It satisfies $\mathbb{E}[w_k w_k^\top] = \mathbf{X}_k^{1/2} \mathbf{U} \mathbf{X}_k^{1/2}$.

We prove the following theorem in this paper for the total regret of FTCL$^{\mathsf{obl}}(T, q, \eta)$.

**Theorem 1.** *In the online eigenvector problem with an* **oblivious** *adversary, there exists absolute constant* $C > 1$ *such that if* $q \geq 3 \log(2dT)$ *and* $\eta \in \left[0, \frac{1}{11q^3}\right]$, *then* FTCL$^{\mathsf{obl}}(T, q, \eta)$ *satisfies*

$$\sum_{k=1}^T \mathbb{E}\left[w_k^\top \mathbf{A}_k w_k\right] = \sum_{k=1}^T \mathbb{E}\left[\mathbf{A}_k \bullet \mathbf{X}_k^{1/2} \mathbf{U} \mathbf{X}_k^{1/2}\right] \geq \bigl(1 - C \cdot \eta q^5 \log(dT)\bigr) \lambda_{\max}(\mathbf{\Sigma}_T) - \frac{4}{\eta} .$$

**Corollary 6.1.** *If* $q = 3 \log(2dT)$ *and* $\eta = \Theta\bigl(\frac{\log^{-3}(dT)}{\sqrt{\lambda_{\max}(\mathbf{\Sigma}_T)}}\bigr)$

$$\sum_{k=1}^T \mathbb{E}\left[w_k^\top \mathbf{A}_k w_k\right] \geq \lambda_{\max}(\mathbf{\Sigma}_T) - O\Bigl(\sqrt{\lambda_{\max}(\mathbf{\Sigma}_T)} \log^3(dT)\Bigr) , \quad (\lambda\text{-refined language})$$

*or choosing the same* $q$ *but* $\eta = \Theta(\log^{-3}(dT)/\sqrt{T})$ *we have*

$$\sum_{k=1}^T \mathbb{E}\left[w_k^\top \mathbf{A}_k w_k\right] \geq \lambda_{\max}(\mathbf{\Sigma}_T) - O\Bigl(\sqrt{T} \log^3(dT)\Bigr) . \quad (\text{general language})$$

As discussed in Section 4, our proof of Theorem 1 relies on a careful analysis on how the potential function $\mathbf{Tr}(\mathbf{X}_k^{1-1/q} \mathbf{U}) = \mathbf{Tr}\bigl((c_k \mathbf{I} - \eta \mathbf{\Sigma}_{k-1})^{-(q-1)} \mathbf{U}\bigr)$ changes across iterations. We analyze this potential increase in two steps: in the first step we replace $\mathbf{\Sigma}_{k-1}$ with $\mathbf{\Sigma}_k$, and in the second step we replace $c_k$ with $c_{k+1}$. After appropriate telescoping, we can derive the result of Theorem 1.

We now discuss the details in the subsequent sections.

---
[14]This $c_k$ is unique because $\mathbf{Tr}(\mathbf{X}_k \mathbf{U})$ is a strictly decreasing function for $c_k > \eta \lambda_{\max}(\mathbf{\Sigma}_{k-1})$.

**Algorithm 1** $\texttt{FTCL}^{\mathsf{obl}}(T, q, \eta)$

**Input:** $T$, number of iterations; $q \geq 2$, an even integer,      ⋄ *theory-predicted choice* $q = \Theta(\log(dT))$
        $\eta$, the learning rate.      ⋄ *theory-predicted choice* $\eta = \log^{-3}(dT)/\sqrt{\lambda_{\max}(\boldsymbol{\Sigma}_T)}$

1: Choose $u_1, u_2, u_3 \in \mathbb{R}^d$ where the $3d$ coordinates are i.i.d. drawn from $\mathcal{N}(0,1)$.
2: $\mathbf{U} \leftarrow \frac{1}{3}\left(u_1 u_1^\top + u_2 u_2^\top + u_3 u_3^\top\right)$.
3: **for** $k \leftarrow 1$ **to** $T$ **do**
4:     $\boldsymbol{\Sigma}_{k-1} \leftarrow \sum_{i=1}^{k-1} \mathbf{A}_i$.
5:     Denote by $\mathbf{X}_k \leftarrow (c_k \mathbf{I} - \eta \boldsymbol{\Sigma}_{k-1})^{-q}$ where $c_k$ is the unique constant satisfying that
       $c_k \mathbf{I} - \eta \boldsymbol{\Sigma}_{k-1} \succ 0$   and   $\mathbf{Tr}(\mathbf{X}_k \mathbf{U}) = 1$ .
6:     Compute $\mathbf{X}_k^{1/2} \mathbf{U} \mathbf{X}_k^{1/2} = \sum_{j=1}^{3} p_j \cdot y_j y_j^\top$ where $y_1, y_2, y_3$ are orthogonal unit vectors in $\mathbb{R}^d$.
7:     Choose $w_k \leftarrow y_j$ with probability $p_j$.     ⋄ *it satisfies* $p_1, p_2, p_3 \geq 0$ *and* $p_1 + p_2 + p_3 = 1$.
8:     Play strategy $w_k$ and receive matrix $\mathbf{A}_k$.
9: **end for**

## 6.1                                                    Well-Behaving Events

Due to concentration reasons, the potential increase could only be "reasonably" bounded for well-behaved matrices $\mathbf{U}$. We now make this definition formal. Given some parameter $\delta > 0$ that we shall later choose to be $1/T^3$, we introduce the following event:

**Definition 6.2.** *For every $k \in \{0, 1, \ldots, T\}$, define event*

$$\mathcal{E}_k(\mathbf{U}) \stackrel{\text{def}}{=} \left\{ \nu_1^\top \mathbf{U} \nu_1 \geq \frac{\delta}{2} \quad \text{and} \quad \forall i \in [d] \colon \nu_i^\top \mathbf{U} \nu_i \leq 2 \log \frac{ed}{\delta} \right\}$$

*where $\nu_1, \ldots, \nu_d$ are the eigenvectors of $\boldsymbol{\Sigma}_k$ with non-increasing eigenvalues. Let $\mathcal{E}_{<j}(\mathbf{U}) \stackrel{\text{def}}{=} \bigwedge_{k=0}^{j-1} \mathcal{E}_k(\mathbf{U})$.*

Intuitively, event $\mathcal{E}_k(\mathbf{U})$ makes sure that the matrix $\mathbf{U}$ is "well-behaved" in the eigenbasis of $\boldsymbol{\Sigma}_k$: (1) it has a non-negligible first coordinate $\nu_1^\top \mathbf{U} \nu_1$, and (2) each coordinate $\nu_i^\top \mathbf{U} \nu_i$ is no more than logarithmic. Using tail bounds for Gaussian distributions, it is not hard to show that this event occurs with probability at least $1 - \delta$ (see Appendix C):

**Lemma 6.3.** $\mathbf{Pr}_{\mathbf{U}}[\mathcal{E}_k(\mathbf{U})] \geq 1 - \delta$ *for all* $k = 0, 1, \ldots, T$.

Under event $\mathcal{E}_{k-1}(\mathbf{U})$, the barrier $c_k$ and the matrix $\mathbf{X}_k$ satisfy the following nice properties. (Their proofs are manipulations of matrix algebra and in Appendix C.)

**Proposition 6.4.** *If $q \geq \max\{\log \frac{2}{\delta}, \log(3d \log \frac{ed}{\delta})\}$, then*

$$\text{event } \mathcal{E}_{k-1}(\mathbf{U}) \text{ implies } \quad \frac{1}{e} \leq c_k - \eta \lambda_{\max}(\boldsymbol{\Sigma}_{k-1}) \leq e \ .$$

*In particular, $\mathcal{E}_{k-1}(\mathbf{U})$ implies (recall $\mathbf{A}_k = \mathbf{P}_k \mathbf{P}_k^\top$)*

*(a)*: $c_k \mathbf{I} - \eta \boldsymbol{\Sigma}_{k-1} \succeq \frac{1}{e} \mathbf{I}$    *(b)*: $\mathbf{Tr}(\mathbf{X}_k^{1-1/q} \mathbf{U}) \leq c_k \leq \eta \lambda_{\max}(\boldsymbol{\Sigma}_{k-1}) + e$    *(c)*: $\eta \mathbf{P}_k^\top \mathbf{X}_k^{1/q} \mathbf{P}_k \preceq e \eta \mathbf{I}$ .

## 6.2                                                    First Potential Increase

The next lemma bounds the potential increase if we replace $\boldsymbol{\Sigma}_{k-1}$ with $\boldsymbol{\Sigma}_k$:

**Lemma 6.5.** *There exists constant $C > 1$ such that, if $q \geq \max\{\log \frac{2}{\delta}, \log(3d \log \frac{ed}{\delta})\}$ and $\eta \leq \frac{1}{3q^3}$,*

$$\mathbb{E}\left[\mathbf{Tr}\left((c_k\mathbf{I} - \eta\mathbf{\Sigma}_k)^{-(q-1)}\mathbf{U}\right) \cdot \mathbb{1}_{\mathcal{E}_{<k}(\mathbf{U})} - \mathbf{Tr}\left((c_k\mathbf{I} - \eta\mathbf{\Sigma}_{k-1})^{-(q-1)}\mathbf{U}\right) \cdot \mathbb{1}_{\mathcal{E}_{<k}(\mathbf{U})}\right]$$
$$\leq (q-1)\eta(1 + C \cdot \eta q^5 \log(d/\delta))\,\mathbb{E}\left[\mathbf{A}_k \bullet \mathbf{X}_k^{1/2}\mathbf{U}\mathbf{X}_k^{1/2}\right] + (\eta T + e)T\delta \ .$$

The proof of Lemma 6.5 is the main technical contribution of this paper, and deviates the most from classical analysis of MMWU. It makes use of our trace inequality in Section 5, and is the only place in our analysis that relies on $\mathbf{rank}(\mathbf{U}) \geq 3$. We include the details in Appendix D.

### 6.3 Second Potential Increase

The following lemma bounds the potential increase if we replace $c_k$ with $c_{k+1}$. Its proof is included in Appendix E and is reasonably straightforward.

**Lemma 6.6.** *For all $q \geq 2$ and $\eta > 0$,*

$$\mathbb{E}\left[\mathbf{Tr}\left((c_{k+1}\mathbf{I} - \eta\mathbf{\Sigma}_k)^{-(q-1)}\mathbf{U}\right) \cdot \mathbb{1}_{\mathcal{E}_{<(k+1)}(\mathbf{U})}\right] - \mathbb{E}\left[\mathbf{Tr}\left((c_k\mathbf{I} - \eta\mathbf{\Sigma}_k)^{-(q-1)}\mathbf{U}\right) \cdot \mathbb{1}_{\mathcal{E}_{<k}(\mathbf{U})}\right]$$
$$\leq -(q-1)(\mathbb{E}[c_{k+1}] - \mathbb{E}[c_k]) \ .$$

Finally, we prove in Appendix F that Theorem 1 is a direct consequence of our two potential increase lemmas above.

## 7 Adversarial Online Eigenvector + Regret in High-Confidence

In this section, we switch to the more challenging adversarial setting: in each iteration $k$, the adversary picks $\mathbf{A}_k$ after seeing the player's strategies $w_1, \ldots, w_{k-1}$. In other words, $\mathbf{A}_k$ may depend on the randomness used in generating $w_1, \ldots, w_{k-1}$ as well.

In such a case, denoting by $\mathcal{D}$ the same rank-3 Wishart distribution we generate $\mathbf{U}$ from in FTCL<sup>obl</sup>, we consider a variant of FTCL<sup>obl</sup> where a new random $\mathbf{U}_k$ is generated from $\mathcal{D}$ per iteration. In other words, instead of choosing $\mathbf{U} \sim \mathcal{D}$ only once at the beginning, we choose $\mathbf{U}_1, \ldots, \mathbf{U}_T$ i.i.d. from $\mathcal{D}$. Then, the normalization constant $c_k$ is defined to satisfy $\mathbf{Tr}((c_k\mathbf{I} - \eta\mathbf{\Sigma}_{k-1})^{-q}\mathbf{U}_k) = 1$. We call this algorithm FTCL<sup>adv</sup> and present it in Algorithm 2 for completeness' sake.

Our next theorem shows that, algorithm FTCL<sup>adv</sup> gives the same regret bound as Theorem 1 even in the adversarial setting; in addition, it elevates the regret bound to a high-confidence level.

**Theorem 2.** *In the online eigenvector problem with an **adversarial** adversary, there exists constant $C > 1$ such that for every $p \in (0, 1)$, $q \geq 3 \log(2dT)$ and $\eta \in \left[0, \frac{1}{11q^3}\right]$, our FTCL<sup>adv</sup>$(T, q, \eta)$ satisfies*

$$w.p. \geq 1 - p: \quad \sum_{k=1}^{T} w_k^\top \mathbf{A}_k w_k \geq \left(1 - C \cdot \eta\big(q^5 \log(dT) + \log(1/p)\big)\right)\lambda_{\max}(\mathbf{\Sigma}_T) - \frac{5}{\eta} \ .$$

**Corollary 7.1.** *Let $q = 3\log(2dT)$ and $\eta = \Theta\big(\frac{\log^3(dT) + \log^{1/2}(1/p))^{-1}}{\sqrt{\lambda_{\max}(\mathbf{\Sigma}_T)}}\big)$, then with prob. $\geq 1 - p$:*

$\sum_{k=1}^{T} w_k^\top \mathbf{A}_k w_k \geq \lambda_{\max}(\mathbf{\Sigma}_T) - \sqrt{\lambda_{\max}(\mathbf{\Sigma}_T)} \cdot O\Big(\log^3(dT) + \sqrt{\log(1/p)}\Big) \ ,$ ($\lambda$-refined language)

*or choosing the same $q$ but $\eta = \Theta\big(\frac{\log^3(dT) + \log^{1/2}(1/p))^{-1}}{\sqrt{T}}\big)$ we have with prob. $\geq 1 - p$:*

$\sum_{k=1}^{T} w_k^\top \mathbf{A}_k w_k \geq \lambda_{\max}(\mathbf{\Sigma}_T) - \sqrt{T} \cdot O\Big(\log^3(dT) + \sqrt{\log(1/p)}\Big) \ .$ (general language)

---

**Algorithm 2** $\texttt{FTCL}^{\textsf{adv}}(T, q, \eta)$

**Input:** $T$, number of iterations;
$\qquad\quad q \geq 2$, an even integer, $\qquad\qquad\qquad\qquad\quad$ ⋄ *theory-predicted choice* $q = \Theta(\log(dT))$
$\qquad\quad \eta$, the learning rate. $\qquad\qquad\qquad\qquad\qquad$ ⋄ *theory-predicted choice* $\eta = \log^{-3}(dT)/\sqrt{\lambda_{\max}(\boldsymbol{\Sigma}_T)}$

1: **for** $k \leftarrow 1$ **to** $T$ **do**
2: $\quad$ Choose 3 vectors $u_1, u_2, u_3 \in \mathbb{R}^d$ where the $3d$ coordinates are i.i.d. drawn from $\mathcal{N}(0,1)$.
3: $\quad \mathbf{U}_k \leftarrow \frac{1}{3}\left(u_1 u_1^\top + u_2 u_2^\top + u_3 u_3^\top\right)$.
4: $\quad \boldsymbol{\Sigma}_{k-1} \leftarrow \sum_{i=1}^{k-1} \mathbf{A}_i$.
5: $\quad$ Denote by $\mathbf{X}_k \leftarrow (c_k \mathbf{I} - \eta \boldsymbol{\Sigma}_{k-1})^{-q}$ where $c_k$ is the unique constant satisfying that
$$c_k \mathbf{I} - \eta \boldsymbol{\Sigma}_{k-1} \succ 0 \quad \text{and} \quad \textbf{Tr}(\mathbf{X}_k \mathbf{U}_k) = 1 \ .$$
6: $\quad$ Compute $\mathbf{X}_k^{1/2} \mathbf{U}_k \mathbf{X}_k^{1/2} = \sum_{j=1}^{3} p_j \cdot y_j y_j^\top$ where $y_1, y_2, y_3$ are orthogonal unit vectors in $\mathbb{R}^d$.
$\qquad$ ⋄ *This is an eigendecomposition and it satisfies $p_1, p_2, p_3 \geq 0$ and $p_1 + p_2 + p_3 = 1$.*
7: $\quad$ Choose $w_k \leftarrow y_j$ with probability $p_j$.
8: $\quad$ Play strategy $w_k$ and receive matrix $\mathbf{A}_k$.
9: **end for**

---

Proof of Theorem 2 relies on a reduction to the oblivious setting, and is included in Appendix G.

## 8 Efficient Implementation of FTCL

Recall that our regret theorems were based on the assumption that in each iteration $k$, the three vectors

$$v_j \stackrel{\text{def}}{=} \mathbf{X}_k^{1/2} u_j = (c_k \mathbf{I} - \eta \boldsymbol{\Sigma}_{k-1})^{-q/2} u_j \quad \text{for } j \in [3] \tag{8.1}$$

can be computed exactly. Once $v_1, v_2, v_3$ are given, we can compute the $3 \times 3$ matrix $(u_i^\top \mathbf{X}_k u_j)_{i,j \in [3]}$ explicitly, from which we can derive in $O(d)$ time the rank-3 eigendecomposition $\mathbf{X}_k^{1/2} \mathbf{U} \mathbf{X}_k^{1/2} = \sum_{j=1}^{3} p_j \cdot y_j y_j^\top$.

Therefore, it suffices to compute $v_1, v_2, v_3$ efficiently. To achieve this goal, we need to

(a) allow $v_1, v_2, v_3$ to be computed approximately,

(b) find the normalization constant $c_k$ efficiently, and

(c) compute $(c_k \mathbf{I} - \eta \boldsymbol{\Sigma}_{k-1})^{-1} b$ efficiently for any $b \in \mathbb{R}^d$.

At a high level, issue (a) is simple because if $v_j'$ satisfies $\|v_j - v_j'\|_2 \leq \widetilde{\varepsilon}/\mathsf{poly}(d, T)$ and we use $v_j'$ instead of $v_j$, then the final regret is affected by less than $\widetilde{\varepsilon}$; issue (b) can be dealt as long as we perform a careful binary search to find $c_k$, similar to prior work [9]; issue (c) can be done as long as we have a good control on the condition number of the matrix $c_k \mathbf{I} - \eta \boldsymbol{\Sigma}_{k-1}$.

We discuss the details in Appendix H, and state below our final running-time theorem:

> **Theorem 3.** *If $q \geq 3\log(2dT/p)$, with probability at least $1 - p$, for all $k \in [T]$, the $k$-th iteration of $\texttt{FTCL}^{\textsf{obl}}$ and $\texttt{FTCL}^{\textsf{adv}}$ runs in $O(d)$ plus the time to solve $\widetilde{O}(1)$ linear systems for matrices $c\mathbf{I} - \eta \boldsymbol{\Sigma}_{k-1}$. Here, $c > 0$ is some constant satisfying $c\mathbf{I} - \eta \boldsymbol{\Sigma}_{k-1} \succ \frac{1}{e} \mathbf{I}$.*

**Corollary 8.1.** *Since the condition number of matrix $c\mathbf{I} - \eta \boldsymbol{\Sigma}_{k-1}$ is at most $\eta \lambda_{\max}(\boldsymbol{\Sigma}_T) + 1$, each linear system can be solved in worst-case time $\widetilde{O}\big(\max\{\sqrt{\eta \lambda_{\max}(\boldsymbol{\Sigma}_T) + 1}, d\} \cdot \mathsf{nnz}(\boldsymbol{\Sigma}_T)\big)$ if implemented by conjugate gradient, or time $\widetilde{O}\big(\mathsf{nnz}(\boldsymbol{\Sigma}_T) + \sqrt{\eta T} \cdot \mathsf{nnz}(\boldsymbol{\Sigma}_T)^{3/4} \mathsf{nnz}(\mathbf{A})^{1/4}\big)$ by the stochastic SVRG method.*

## 9 Stochastic Online Eigenvector

Consider the special case when the matrices $\mathbf{A}_1, \ldots, \mathbf{A}_T$ are generated i.i.d. from a common distribution whose expectation equals $\mathbf{B}$. This is known as the stochastic online eigenvector problem, and we wish to minimize the regret[15]

$$\sum_{k=1}^T w_k^\top \mathbf{A}_k w_k - T \cdot \lambda_{\max}(\mathbf{B}) \ .$$

We revisit Oja's algorithm: beginning with a random Gaussian vector $u \in \mathbb{R}^d$, at each iteration $k$, let $w_k$ be $(\mathbf{I} + \eta \mathbf{A}_{k-1}) \cdots (\mathbf{I} + \eta \mathbf{A}_1) u$ after normalization. It is clear that $w_k$ can be computed from $w_{k-1}$ in time $\mathsf{nnz}(\mathbf{A})$.

We include in Appendix I a one-paged proof of the following theorem:

**Theorem 4.** *There exists $C > 1$ such that, for every $p \in (0,1)$, if $\eta \in \big[0, \sqrt{p/(75T\lambda_{\max}(\mathbf{B}))}\big]$ in Oja's algorithm, we have with probability at least $1 - p$:*

$$\sum_{k=1}^T w_k^\top \mathbf{A}_k w_k \geq (1 - 2\eta) T \cdot \lambda_{\max}(\mathbf{B}) - C \cdot \frac{\log(d/p)}{\eta} \ .$$

**Corollary 9.1.** *Choosing $\eta = \Theta(\sqrt{p}/\sqrt{T\lambda_{\max}(\mathbf{B})})$, we have with prob. $\geq 1 - p$:*

$$\sum_{k=1}^T w_k^\top \mathbf{A}_k w_k \geq T \cdot \lambda_{\max}(\mathbf{B}) - O\big(\frac{\sqrt{T \cdot \lambda_{\max}(\mathbf{B})}}{\sqrt{p}} \cdot \log(d/p)\big) \ . \qquad (\lambda\text{-refined language})$$

*Choosing $\eta = \Theta(\sqrt{p}/\sqrt{T})$, we have with prob. $\geq 1 - p$:*

$$\sum_{k=1}^T w_k^\top \mathbf{A}_k w_k \geq T \cdot \lambda_{\max}(\mathbf{B}) - O\big(\frac{\sqrt{T}}{\sqrt{p}} \cdot \log(d/p)\big) \ . \qquad (\text{general language})$$

The proof of Theorem 4 uses a potential function analysis which is similar to the matrix exponential potential used in MMWU, but compressed to dimension 1.

## 10 Conclusions

We give a new learning algorithm FTCL for the online eigenvector problem. It matches the optimum regret obtained by MMWU, but runs *much faster*. It matches the fast per-iteration running time of FTPL, but has a *much smaller regret*. In the stochastic setting, our side result on Oja's algorithm also outperforms previous results. We believe our novel idea of "follow the compressed leader" may find other applications in the future.

## Acknowledgement

We thank Yin Tat Lee for discussing the problem regarding how to compress MMWU to constant dimension in 2015. We thank Elad Hazan for suggesting us the problem and for several insightful discussions. We thank Dan Garber and Tengyu Ma for clarifying some results of prior work [20].

---

[15]In principle, one can also ask to minimize regret where $T \cdot \mathbf{B}$ is replaced with $\mathbf{A}_1 + \cdots + \mathbf{A}_T$. However, due to simple concentration results, there is no big difference between the two different notions. [20]

# Appendix

| Paper | Total Regret | Time Per Iteration | Minimum Total Time for $\varepsilon$ Average Regret |
|---|---|---|---|
| MMWU [9, 11] | $\widetilde{O}(\sqrt{\lambda T})$ | at least $O(d^\omega)$ | $\widetilde{O}\big(\frac{\lambda d^\omega}{\varepsilon^2}\big)$ |
| MMWU-JL [9, 35] | $\widetilde{O}(\sqrt{\lambda T})$ | $\mathsf{M}^{\mathsf{exp}} \times \widetilde{O}(T/\lambda)$ | $\widetilde{O}\big(\frac{\lambda^{1.5}}{\varepsilon^{4.5}}\mathsf{nnz}(\boldsymbol{\Sigma})\big)$ |
| FTPL [20] | not clear | | |
| **this paper** | $\widetilde{O}(\sqrt{\lambda T})$ Theorem 1&2 | $\mathsf{M}^{\mathsf{lin}} \times \widetilde{O}(1)$ Theorem 3 | $\widetilde{O}\big(\frac{\lambda^{1.5}}{\varepsilon^{2.5}}\mathsf{nnz}(\boldsymbol{\Sigma})\big)$ and $\widetilde{O}\big(\frac{\lambda}{\varepsilon^{2.5}}\mathsf{nnz}(\boldsymbol{\Sigma})^{\frac{3}{4}}\mathsf{nnz}(\mathbf{A})^{\frac{1}{4}} + \frac{\lambda}{\varepsilon^2}\mathsf{nnz}(\boldsymbol{\Sigma})\big)$ |
| ↓ stochastic online eigenvector only ↓ | | | |
| block power method [20] | $\widetilde{O}(\sqrt{\lambda T})$ | $O(\mathsf{nnz}(\boldsymbol{\Sigma}))$ | $\widetilde{O}\big(\frac{\lambda}{\varepsilon^2}\mathsf{nnz}(\boldsymbol{\Sigma})\big)$ |
| **this paper** | $\widetilde{O}(\sqrt{\lambda T})$ Theorem 4 | $O(\mathsf{nnz}(\mathbf{A}))$ Theorem 4 | $\widetilde{O}\big(\frac{\lambda}{\varepsilon^2}\mathsf{nnz}(\mathbf{A})\big)$ |

Table 2: Comparison of known methods for the online eigenvector problem in the $\lambda$-refined language (see Section 1.4). We denote by $\boldsymbol{\Sigma} = \mathbf{A}_1 + \cdots + \mathbf{A}_T$, by $\mathsf{nnz}(\mathbf{A}) = \max_{k \in [T]} \{\mathsf{nnz}(\mathbf{A}_k)\}$, and by $\lambda = \frac{1}{T}\lambda_{\max}(\boldsymbol{\Sigma}) \in [0, 1]$.
- $\mathsf{M}^{\mathsf{exp}}$ is the time to compute $e^{-\mathbf{M}}$ multiplied with a vector, where $\mathbf{M} \in \mathbb{R}^{d \times d}$ satisfies $0 \preceq \mathbf{M} \preceq \widetilde{O}((\lambda T)^{1/2}) \cdot \mathbf{I}$.
- $\mathsf{M}^{\mathsf{lin}}$ is the time to solve a linear system in $\mathbf{M} \in \mathbb{R}^{d \times d}$, where $\mathbf{M}$ is PSD and of condition number $\leq \widetilde{O}((\lambda T)^{1/2})$.
- If using iterative methods, the worst-case values $\mathsf{M}^{\mathsf{exp}}$ and $\mathsf{M}^{\mathsf{lin}}$ are

$$\mathsf{M}^{\mathsf{exp}} = \widetilde{O}\big(\min\{(\lambda T)^{\frac{1}{4}}\mathsf{nnz}(\boldsymbol{\Sigma}), d^\omega\}\big) \geq \mathsf{M}^{\mathsf{lin}} = \widetilde{O}\big(\min\{\min\{d, (\lambda T)^{\frac{1}{4}}\}\mathsf{nnz}(\boldsymbol{\Sigma}), d^\omega\}\big) \enspace,$$

where $d^\omega$ is the time needed to multiply two $d \times d$ matrices. If using stochastic iterative methods, $\mathsf{M}^{\mathsf{lin}}$ is at most $\widetilde{O}\big((T/\lambda)^{\frac{1}{4}}\mathsf{nnz}(\boldsymbol{\Sigma})^{\frac{3}{4}}\mathsf{nnz}(\mathbf{A})^{\frac{1}{4}} + \mathsf{nnz}(\boldsymbol{\Sigma})\big)$. (See discussions in Section 3.)

## A  Evaluation Setup

Recall that our FTCL has nearly-optimal $\widetilde{O}(\sqrt{T})$ total regret, just like MMWU or MMWU-JL. However, the previous developed FTPL method has a total regret $\widetilde{O}(\sqrt{dT})$ and this could be far from optimal. In this section, we generate synthetic data to verify that FTPL can indeed have poor regret performance.

We generate three sequences of synthetic matrices $\mathbf{A}_k$:

1. `random`. We pick a random covariance matrix $\boldsymbol{\Sigma}$ from Wishart distribution. In each iteration $k$, we pick a random vector $v_k \sim \mathcal{N}(0, \boldsymbol{\Sigma})$ and let $\mathbf{A}_k = v_k v_k^\top$. Note that matrices $\mathbf{A}_k$ are i.i.d.

2. `diagonal`. In iteration $k$ where $k = \frac{sd}{2} + r$ for $s \in \mathbb{N}$ and $r \in [d/2]$, we whose $\mathbf{A}_k = \frac{1}{2}\mathbf{I} + \mathbf{E}_r$, where $\mathbf{E}_r$ a matrix with all entries zero except the $(r, r)$ entry being 1. In dataset `diagonal`, the eigen basis is fixed, and each vector $e_i$ in the standard basis takes turns to be the leading eigenvector.

3. `diagonal+rotation`. In iteratin $k$ where $k = \frac{sd}{2} + r$, we whose $\mathbf{A}_k = \mathbf{P}^\top(\frac{1}{2}\mathbf{I} + \mathbf{E}_r)\mathbf{P}$, where $\mathbf{P}_s$ is a "rotation matrix" whose entries are

$$[\mathbf{P}_s]_{i,j} = \begin{cases} \cos(\frac{4\pi sd}{T}) & \text{if } i = j; \\ \sin(\frac{4\pi sd}{T}) & \text{if } i = j - 1 \\ -\sin(\frac{4\pi sd}{T}) & \text{if } i = j + 1 \\ 0 & \text{otherwise.} \end{cases}$$

This dataset `diagonal+rotation` is just dataset `diagonal` plus a rotation in each step, so the eigen basis of the matrix gradually changes.

We pick dimension $d = 100$ and $T = 10000$, and have implemented FTPL, FTCL and MMWU. (We did not implement MMWU-JL because MMWU has better regret than MMWU-JL.) For each of the three algorithms, we search through 100 different parameters for the learning rate, and report the best total regret.

As illustrated in Figure 1, We can see that when the matrices are random, three algorithms behaves similarly. However, even in the simple data where each diagonal entries keep turns to be large, our algorithm has a notable advantage over FTPL. When the eigen basis starts to change, FTPL behaves significantly worse than FTCL and MMWU.

## B  Proof of Lemma 5.1

We first recall Markov brother's inequality. For a polynomial $f : \mathbb{R} \to \mathbb{R}$, we use $f^{(k)}$ to denote the $k$-th order derivative of $f$ at point $x$. We have:

**Theorem B.1** (Markov brother's inequality)**.** *If polynomial $f$ is of degree $n$, then $\forall k \in \mathbb{N}^*$ and $\forall a > 0$:*

$$\max_{x \in [0,a]} |f^{(k)}(x)| \leq \left(\frac{2}{a}\right)^i \frac{n^2(n^2 - 1^2)(n^2 - 2^2)\ldots(n^2 - (k-1)^2)}{(2k-1)!!} \max_{x \in [0,a]} |f(x)| \ . \qquad \text{(B.1)}$$

**Lemma 5.1.** *For every symmetric matrices $\mathbf{A}, \mathbf{B}, \mathbf{D} \in \mathbb{R}^{d \times d}$, every integer $k \geq 1$, every $\eta^* \geq 0$, and every $\eta \in [0, \eta^*/k^2]$, if $\mathbf{A}$ and $\mathbf{D}$ are* commutative, *then*

$$(\mathbf{A} + \eta\mathbf{B})^k \bullet \mathbf{D} - \mathbf{A}^k \bullet \mathbf{D} \leq k\eta\mathbf{B} \bullet \mathbf{A}^{k-1}\mathbf{D} + \left(\frac{\eta k^2}{\eta^*}\right)^2 \max_{\eta' \in [0,\eta^*]} \left\{\left|(\mathbf{A} + \eta'\mathbf{B})^k \bullet \mathbf{D} - \mathbf{A}^k \bullet \mathbf{D}\right|\right\} \ .$$

*Proof.* Consider a degree-$k$ polynomial

$$f(\eta) \stackrel{\text{def}}{=} (\mathbf{A} + \eta\mathbf{B})^k \bullet \mathbf{D} - \mathbf{A}^k \bullet \mathbf{D} = \sum_{i=1}^{k} \eta^i \sum_{\substack{j_0, \ldots, j_i \in \mathbb{Z}_{\geq 0} \\ j_0 + \cdots + j_i = k-i}} \mathbf{A}^{j_0}\mathbf{B}\mathbf{A}^{j_1}\mathbf{B}\cdots\mathbf{B}\mathbf{A}^{j_i} \bullet \mathbf{D}$$

Its first order derivative

$$f'(0) = \sum_{\substack{j_0, j_1 \in \mathbb{Z}_{\geq 0} \\ j_0 + j_1 = k-1}} \mathbf{A}^{j_0}\mathbf{B}\mathbf{A}^{j_1} \bullet \mathbf{D} = \sum_{\substack{j_0, j_1 \in \mathbb{Z}_{\geq 0} \\ j_0 + j_1 = k-1}} \mathbf{A}^{(k-1)/2}\mathbf{B}\mathbf{A}^{(k-1)/2} \bullet \mathbf{D} = k\mathbf{B} \bullet \mathbf{A}^{(k-1)/2}\mathbf{D}\mathbf{A}^{(k-1)/2} \ .$$

Above, the first equality is due to the commutativity between $\mathbf{A}$ and $\mathbf{D}$. Letting $f^* \stackrel{\text{def}}{=} \max_{\eta' \in [0,\eta^*]} |f(\eta')|$, we can apply Markov brothers' inequality (B.1) and obtain for every $i \geq 2$,

$$|f^{(i)}(0)| \leq \left(\frac{2}{\eta^*}\right)^i \cdot \frac{k^2(k^2-1)\cdots(k^2-(i-1)^2)}{1 \cdot 3 \cdot 5 \cdots (2i-1)} \max_{\eta' \in [0,\eta^*]} |f(\eta')| \leq \frac{k^{2i}}{(\eta^*)^i} f^* \ .$$

Therefore, as long as $\eta \leq \frac{\eta^*}{k^2}$, we have

$$f(\eta) = f(0) + f'(0) \cdot \eta + \sum_{i=2}^{k} \eta^i \cdot \frac{f^{(i)}(0)}{i!} \leq f(0) + f'(0) \cdot \eta + \sum_{i=2}^{k} \left(\frac{\eta k^2}{\eta^*}\right)^i \cdot \frac{f^*}{i!}$$

$$\leq f(0) + f'(0) \cdot \eta + \left(\frac{\eta k^2}{\eta^*}\right)^2 f^* \ .$$

Since $f(0) = 0$ we complete the proof. □

# C Proof for Section 6.1

## C.1 Proof of Lemma 6.3

**Lemma 6.3.** *For every $k = 0, 1, \ldots, T$, we have $\mathbf{Pr}_{\mathbf{U}}[\mathcal{E}_k(\mathbf{U})] \geq 1 - \delta$.*

*Proof.* Let $\nu_1, \ldots, \nu_d$ be the eigenvectors of $\boldsymbol{\Sigma}_k$ with non-increasing eigenvalues. Because Gaussian random vectors are rotationally invariant, we can view each $u_1, u_2, u_3$ as drawn in the basis of $\nu_1, \ldots, \nu_d$, so each $\nu_i^\top u_j$ is drawn i.i.d. from $\mathcal{N}(0, 1)$ for every $i \in [d], j \in [3]$.

Since $\nu_1^\top \mathbf{U} \nu_1 = \frac{1}{3}\big((\nu_1^\top u_1)^2 + (\nu_1^\top u_2)^2 + (\nu_1^\top u_3)^2\big)$, we immediately know that $3\nu_1^\top \mathbf{U} \nu_1$ is distributed according to chi-square distribution $\chi^2(3)$. The probability density function of $\chi^2(3)$ is $f(x) = \frac{e^{-x/2}\sqrt{x}}{\sqrt{2\pi}}$ (for $x \in [0, \infty)$) and therefore

$$\mathbf{Pr}\left[\nu_1^\top \mathbf{U}\nu_1 \leq \delta/2\right] \leq \int_0^{3\delta/2} \frac{e^{-x/2}\sqrt{x}}{\sqrt{2\pi}} dx \leq \int_0^{3\delta/2} \frac{\sqrt{x}}{\sqrt{2\pi}} dx = \frac{1}{2}\sqrt{\frac{3}{\pi}}\delta^{3/2} \leq \frac{\delta}{2} \ .$$

As for the second condition, for every $t \geq 0$ and $i \in [d]$,

$$\mathbf{Pr}\left[\nu_i^\top \mathbf{U}\nu_i \geq t/3\right] \leq \int_t^\infty \frac{e^{-x/2}\sqrt{x}}{\sqrt{2\pi}} dx = 1 - \mathrm{Erf}\left(\frac{\sqrt{t}}{\sqrt{2}}\right) + \sqrt{\frac{2}{\pi}} e^{-t/2}\sqrt{t} \leq e^{-t/2} + \sqrt{\frac{2}{\pi}} e^{-t/2}\sqrt{t} \ ,$$

where $\mathrm{Erf}(x)$ is the Gauss error function. Picking $t = 4 \log \frac{ed}{\delta}$, we have

$$e^{-t/2} + \sqrt{\frac{2}{\pi}} e^{-t/2}\sqrt{t} \leq \frac{\delta^2}{e^2 d^2} + \sqrt{\frac{2}{\pi}} \frac{\delta^2}{e^2 d^2} \cdot 2\sqrt{\frac{ed}{\delta}} < \frac{\delta}{2d} \ .$$

Therefore, we have $\mathbf{Pr}\left[\forall i \in [d] \colon \nu_i^\top \mathbf{U}\nu_i \geq 2 \log \frac{ed}{\delta}\right] \leq \frac{\delta}{2}$ and we conclude by union bound $\mathbf{Pr}_{\mathbf{U}}[\overline{\mathcal{E}_k(\mathbf{U})}] \leq \frac{\delta}{2} + \frac{\delta}{2} = \delta$ . $\square$

## C.2 Proof of Proposition 6.4

**Proposition 6.4.** *If $q \geq \max\{\log \frac{2}{\delta}, \log(3d \log \frac{ed}{\delta})\}$, then*

$$\textit{event } \mathcal{E}_{k-1}(\mathbf{U}) \textit{ implies } \quad \frac{1}{e} \leq c_k - \eta \lambda_{\max}(\boldsymbol{\Sigma}_{k-1}) \leq e \ . \tag{C.1}$$

*In particular, $\mathcal{E}_{k-1}(\mathbf{U})$ implies (recall $\mathbf{A}_k = \mathbf{P}_k \mathbf{P}_k^\top$)*

(a): $c_k \mathbf{I} - \eta \boldsymbol{\Sigma}_{k-1} \succeq \frac{1}{e} \mathbf{I}$    (b): $\mathbf{Tr}(\mathbf{X}_k^{1-1/q} \mathbf{U}) \leq c_k \leq \eta \lambda_{\max}(\boldsymbol{\Sigma}_{k-1}) + e$    (c): $\eta \mathbf{P}_k^\top \mathbf{X}_k^{1/q} \mathbf{P}_k \preceq e\eta \mathbf{I}$ .

*Proof.* Let $\nu_1, \ldots, \nu_d$ be the eigenvectors of $\boldsymbol{\Sigma}_{k-1}$ with non-increasing eigenvalues $\lambda_1, \ldots, \lambda_d$. Then, $\sum_{i=1}^d \frac{\nu_i^\top \mathbf{U} \nu_i}{(c_k - \eta \lambda_i)^q} = \mathbf{Tr}(\mathbf{X}_k \mathbf{U}) = 1$. However, event $\mathcal{E}_k(\mathbf{U})$ tells us $\nu_i^\top \mathbf{U} \nu_i \geq \frac{\delta}{2}$ which implies $(c_k - \eta \lambda_1)^q \geq \frac{\delta}{2}$. Under our choice of $q$, we have $c_k - \eta \lambda_1 \geq \frac{1}{e}$ which proves the first inequality in (C.1).

On the other hand, letting $c = \eta \lambda_{\max}(\boldsymbol{\Sigma}_{k-1}) + e$, our choice of $q$ implies

$$\mathbf{Tr}((c\mathbf{I} - \eta \boldsymbol{\Sigma}_{k-1})^{-q} \mathbf{U}) = \sum_{i=1}^d \frac{\nu_i^\top \mathbf{U} \nu_i}{(c - \eta \lambda_i)^q} \leq \sum_{i=1}^d \frac{2\log(ed/\delta)}{e^q} \leq 1 \ .$$

Since the left hand side of the above inequality is an decreasing function in $c$, and since $\mathbf{Tr}((c_k \mathbf{I} - \eta \boldsymbol{\Sigma}_{k-1})^{-q} \mathbf{U}) = 1$, we must have $c_k \leq c$ which proves the second inequality in (C.1).

Finally, (a) is a simple corollary of the first inequality of (C.1). As for (b), it simply comes from the following upper bound

$$\mathbf{Tr}(\mathbf{X}_k^{1-1/q} \mathbf{U}) = \sum_{i=1}^d \frac{\nu_i^\top \mathbf{U}\nu_i}{(c_k - \eta \lambda_i)^{q-1}} \leq c_k \sum_{i=1}^d \frac{\nu_i^\top \mathbf{U}\nu_i}{(c_k - \eta \lambda_i)^q} = c_k \mathbf{Tr}(\mathbf{X}_k \mathbf{U}) = c_k \ .$$

As for (c), it follows from $\mathbf{P}_k^\top \mathbf{P}_k \preceq \mathbf{I}$ so $\eta \mathbf{P}_k^\top \mathbf{X}_k^{1/q} \mathbf{P}_k \preceq \eta \mathbf{P}_k^\top (e\mathbf{I}) \mathbf{P}_k \preceq e\eta \mathbf{I}$. □

## D  Proof for Section 6.2

**Lemma 6.5.** *There is constant $C > 1$ such that, if $q \geq \max\{\log \frac{2}{\delta}, \log(3d \log \frac{ed}{\delta})\}$ and $\eta \leq \frac{1}{11q^3}$,*

$$\mathbb{E}\left[\mathbf{Tr}\left((c_k\mathbf{I} - \eta\mathbf{\Sigma}_k)^{-(q-1)}\mathbf{U}\right) \cdot \mathbb{1}_{\mathcal{E}_{<k}(\mathbf{U})} - \mathbf{Tr}\left((c_k\mathbf{I} - \eta\mathbf{\Sigma}_{k-1})^{-(q-1)}\mathbf{U}\right) \cdot \mathbb{1}_{\mathcal{E}_{<k}(\mathbf{U})}\right]$$
$$\leq (q-1)\eta(1 + C \cdot \eta q^5 \log(d/\delta))\, \mathbb{E}\left[\mathbf{A}_k \bullet \mathbf{X}_k^{1/2} \mathbf{U} \mathbf{X}_k^{1/2}\right] + (\eta T + e)T\delta \ .$$

*Remark* D.1. We have slightly abused notations here. In principle, the quantity $\mathbf{Tr}\left((c_k\mathbf{I} - \eta\mathbf{\Sigma}_k)^{-(q-1)}\mathbf{U}\right)$ can be unbounded if $c_k\mathbf{I} - \eta\mathbf{\Sigma}_k$ is not invertible. However, as we shall see in the proof of Lemma 6.5, this necessarily implies $\mathbb{1}_{\mathcal{E}_{<k}(\mathbf{U})} = 0$ because of Proposition 6.4. Therefore, we define $\mathbf{Tr}\left((c_k\mathbf{I} - \eta\mathbf{\Sigma}_k)^{-(q-1)}\mathbf{U}\right) \cdot \mathbb{1}_{\mathcal{E}_{<k}(\mathbf{U})}$ to be zero if this happens.

*Proof of Lemma 6.5.* Let $\nu_1, \ldots, \nu_d$ be the eigenvectors of $\mathbf{\Sigma}_{k-1}$ with non-increasing eigenvalues. In this proof, let us assume without loss of generality that all vectors and matrices are written in this eigenbasis (so $\mathbf{\Sigma}_{k-1}$ and $\mathbf{X}_k$ are both diagonal matrix).

Since Gaussian random vectors are rotationally invariant, we assume that $u_1, u_2, u_3$ are generated according to the following procedure: first, the absolute values of their $3d$ coordinates $u_1, u_2, u_3$ are determined; then, their signs are determined.

Denoting by $\mathbf{D} = \mathsf{diag}\{\mathbf{U}_{11}, \ldots, \mathbf{U}_{dd}\}$ the diagonal part of $\mathbf{U}$, we immediately notice that $\mathbf{D}$ is determined *completely* at the first step of the above procedure. This has two important consequences that we shall rely crucially in the proof:

- fixing the randomness of $\mathbf{D}$, it satisfies $\mathbb{E}_\mathbf{U}[\mathbf{U}|\mathbf{D}] = \mathbf{D}$;[16]
- $c_k$ is completely determined by $\mathbf{D}$. [17]

In addition, since the event $\mathcal{E}_{k-1}(\mathbf{U})$ only depends on the diagonal entry of $\mathbf{U}$, slightly abusing notation, we also use $\mathcal{E}_{k-1}(\mathbf{D})$ to denote this event on diagonal matrices $\mathbf{D}$. We also use $D_i$ to represent the $i$-th diagonal entry of $\mathbf{D}$. Our proof now has three parts:

**Part I: Potential Increase for D.**  For every PSD matrix $\mathbf{D}$, denoting by $\mathbf{A}_k = \mathbf{P}_k \mathbf{P}_k^\top$,

$$\mathbf{Tr}\left((c_k\mathbf{I} - \eta\mathbf{\Sigma}_k)^{-(q-1)}\mathbf{D}\right) - \mathbf{Tr}\left((c_k\mathbf{I} - \eta\mathbf{\Sigma}_{k-1})^{-(q-1)}\mathbf{D}\right)$$
$$\overset{①}{=} \mathbf{Tr}\left((\mathbf{X}_k^{-1/q} - \eta\mathbf{P}_k\mathbf{P}_k^\top)^{-(q-1)}\mathbf{D}\right) - \mathbf{Tr}\left(\mathbf{X}_k^{1-1/q}\mathbf{D}\right)$$
$$\overset{②}{=} \mathbf{Tr}\left(\left(\mathbf{X}_k^{1/q} + \eta\mathbf{X}_k^{1/q}\mathbf{P}_k(\mathbf{I} - \eta\mathbf{P}_k^\top \mathbf{X}_k^{1/q}\mathbf{P}_k)^{-1}\mathbf{P}_k^\top \mathbf{X}_k^{1/q}\right)^{q-1}\mathbf{D}\right) - \mathbf{Tr}\left(\mathbf{X}_k^{1-1/q}\mathbf{D}\right) \quad \text{(D.1)}$$

Above, ① follows from the definition of $\mathbf{X}_k$ and ② uses the Woodbury formula for matrix inversion.

Now, unlike the classical proof for MMWU, our matrix $\mathbf{D}$ here is *not* identity so we cannot rely on the Lieb-Thirring trace inequality to bound the right hande side of (D.1) like it was used in [9]. We can instead consult our new trace inequality Lemma 5.1 because $\mathbf{D}$ and $\mathbf{X}_k$ are both diagonal matrices so they are commutative. Recall that Lemma 5.1 requires a crude upper bound on the first trace quantity on the term "$|(\mathbf{A} + \eta'\mathbf{B})^k \bullet \mathbf{D} - \mathbf{A}^k \bullet \mathbf{D}|$", and we shall provide this crude upper bound in Lemma D.2.

---
[16] More specifically, since the off-diagonal entries of $\mathbf{U}$ can still randomly flip signs in the second step of the random procedure, their expectations are all equal to zero.

[17] This is because $c_k$ is defined as the constant satisfying $1 = \mathbf{Tr}((c_k\mathbf{I} - \eta\mathbf{\Sigma}_{k-1})\mathbf{U}) = \mathbf{Tr}((c_k\mathbf{I} - \eta\mathbf{\Sigma}_{k-1})\mathbf{D})$.

Formally, choosing $\eta^* \stackrel{\text{def}}{=} \frac{1}{11q}$, we that for every $\mathbf{D}$ satisfying $\mathcal{E}_{k-1}(\mathbf{D})$,

$$\mathbf{Tr}\Big((c_k\mathbf{I} - \eta\mathbf{\Sigma}_k)^{-(q-1)}\mathbf{D}\Big) - \mathbf{Tr}\Big((c_k\mathbf{I} - \eta\mathbf{\Sigma}_{k-1})^{-(q-1)}\mathbf{D}\Big)$$

$$\stackrel{\text{③}}{\leq} (q-1)\eta \mathbf{X}_k^{1/q}\mathbf{P}_k(\mathbf{I} - \eta\mathbf{P}_k^\top\mathbf{X}_k^{1/q}\mathbf{P}_k)^{-1}\mathbf{P}_k^\top\mathbf{X}_k^{1/q} \bullet \mathbf{X}_k^{(q-2)/q}\mathbf{D}$$

$$+ \left(\frac{\eta(q-1)^2}{\eta^*}\right)^2 \cdot 4(q-1)\eta^* \|\mathbf{D}\|_2 \mathbf{P}_k\mathbf{P}_k^\top \bullet \mathbf{X}_k$$

$$\stackrel{\text{④}}{\leq} \frac{(q-1)\eta}{1-e\eta}\mathbf{P}_k\mathbf{P}_k^\top \bullet \mathbf{X}_k\mathbf{D} + O(\eta^2 q^6) \cdot \|\mathbf{D}\|_2 \cdot \mathbf{P}_k\mathbf{P}_k^\top \bullet \mathbf{X}_k$$

$$\stackrel{\text{⑤}}{\leq} (q-1)\eta \mathbf{P}_k\mathbf{P}_k^\top \bullet \mathbf{X}_k\mathbf{D} + O(\eta^2 q^6 \log(d/\delta))\mathbf{P}_k\mathbf{P}_k^\top \bullet \mathbf{X}_k \ . \tag{D.2}$$

Above, ③ follows from Lemma 5.1 (with $\eta \leq \eta^*/q^2$) together with Lemma D.2 (for $\eta = \eta^*$); ④ follows from $\mathbf{I} - \eta\mathbf{P}_k^\top\mathbf{X}_k^{1/q}\mathbf{P}_k \succeq (1 - e\eta)\mathbf{I}$ (see Proposition 6.4), the fact that $\mathbf{Tr}(\mathbf{AC}) \leq \mathbf{Tr}(\mathbf{BC})$ for $\mathbf{A} \preceq \mathbf{B}$ and $\mathbf{C}$ symmetric, and the choice of $\eta^*$; ⑤ follows from our assumption $\eta \leq \frac{1}{6}$ as well as $\|\mathbf{D}\|_2 \leq 2\log\frac{2d}{\delta}$ which comes from the definition of event $\mathcal{E}_{k-1}(\mathbf{D})$.

**Part II: Potential Increase for All U That Agrees With D.** For every fixed $\mathbf{D}$ that satisfies $\mathcal{E}_{k-1}(\mathbf{D})$, taking expectation over all matrices $\mathbf{U}$ that agrees with $\mathbf{D}$:[18]

$$\mathbb{E}\Big[\mathbf{Tr}\Big((c_k\mathbf{I} - \eta\mathbf{\Sigma}_k)^{-(q-1)}\mathbf{U}\Big) \cdot \mathbb{1}_{\mathcal{E}_{<(k-1)}(\mathbf{U})} - \mathbf{Tr}\Big((c_k\mathbf{I} - \eta\mathbf{\Sigma}_{k-1})^{-(q-1)}\mathbf{U}\Big) \cdot \mathbb{1}_{\mathcal{E}_{<(k-1)}(\mathbf{U})}\Big|\mathbf{D}\Big]$$

$$\stackrel{\text{①}}{\leq} \mathbb{E}\Big[\mathbf{Tr}\Big((c_k\mathbf{I} - \eta\mathbf{\Sigma}_k)^{-(q-1)}\mathbf{U}\Big) - \mathbf{Tr}\Big((c_k\mathbf{I} - \eta\mathbf{\Sigma}_{k-1})^{-(q-1)}\mathbf{U}\Big)$$

$$+ \mathbf{Tr}\Big((c_k\mathbf{I} - \eta\mathbf{\Sigma}_{k-1})^{-(q-1)}\mathbf{U}\Big) \cdot (1 - \mathbb{1}_{\mathcal{E}_{<(k-1)}(\mathbf{U})})\Big|\mathbf{D}\Big]$$

$$\stackrel{\text{②}}{\leq} \mathbf{Tr}\Big((c_k\mathbf{I} - \eta\mathbf{\Sigma}_k)^{-(q-1)}\mathbf{D}\Big) - \mathbf{Tr}\Big((c_k\mathbf{I} - \eta\mathbf{\Sigma}_{k-1})^{-(q-1)}\mathbf{D}\Big) + \mathbb{E}\Big[(\eta T + e) \cdot (1 - \mathbb{1}_{\mathcal{E}_{<(k-1)}(\mathbf{U})})\Big|\mathbf{D}\Big]$$

$$= \mathbf{Tr}\Big((c_k\mathbf{I} - \eta\mathbf{\Sigma}_k)^{-(q-1)}\mathbf{D}\Big) - \mathbf{Tr}\Big((c_k\mathbf{I} - \eta\mathbf{\Sigma}_{k-1})^{-(q-1)}\mathbf{D}\Big) + (\eta T + e) \cdot \mathbf{Pr}\Big[\overline{\mathcal{E}_{<(k-1)}(\mathbf{U})}\Big|\mathbf{D}\Big]$$

$$\stackrel{\text{③}}{\leq} (q-1)\eta\mathbf{P}_k\mathbf{P}_k^\top \bullet \mathbf{X}_k\mathbf{D} + O(\eta^2 q^6 \log(d/\delta)) \cdot \mathbf{P}_k\mathbf{P}_k^\top \bullet \mathbf{X}_k + (\eta T + e)T\delta$$

$$\stackrel{\text{④}}{=} (q-1)\eta \, \mathbb{E}\big[\mathbf{P}_k\mathbf{P}_k^\top \bullet \mathbf{X}_k\mathbf{U} \mid \mathbf{D}\big] + O(\eta^2 q^6 \log(d/\delta)) \cdot \mathbf{P}_k\mathbf{P}_k^\top \bullet \mathbf{X}_k + (\eta T + e)T\delta. \tag{D.3}$$

Above, ① is because indicator functions are never greater than 1; ② uses $\mathbf{Tr}(\mathbf{X}_k^{1-1/q}\mathbf{U}) \leq \eta\lambda_{\max}(\mathbf{\Sigma}_{k-1}) + e \leq \eta T + e$ which follows from Proposition 6.4; ③ follows from (D.2) as well as Lemma 6.3; and ④ follows from the observation $\mathbb{E}_\mathbf{U}[\mathbf{U}|\mathbf{D}] = \mathbf{D}$ together with the fact that $\mathbf{X}_k$ only depends on $\mathbf{D}$.

**Part III: Potential Increase for All U.** We now claim for *all* possible diagonal $\mathbf{D}$, it satisfies

$$\mathbb{E}\Big[\mathbf{Tr}\Big((c_k\mathbf{I} - \eta\mathbf{\Sigma}_k)^{-(q-1)}\mathbf{U}\Big) \cdot \mathbb{1}_{\mathcal{E}_{<k}(\mathbf{U})} - \mathbf{Tr}\Big((c_k\mathbf{I} - \eta\mathbf{\Sigma}_{k-1})^{-(q-1)}\mathbf{U}\Big) \cdot \mathbb{1}_{\mathcal{E}_{<k}(\mathbf{U})}\Big|\mathbf{D}\Big]$$

$$\leq (q-1)\eta \, \mathbb{E}\big[\mathbf{P}_k\mathbf{P}_k^\top \bullet \mathbf{X}_k\mathbf{U}|\mathbf{D}\big] + O(\eta^2 q^6 \log(d/\delta)) \cdot \mathbf{P}_k\mathbf{P}_k^\top \bullet \mathbf{X}_k + (\eta T + e)T\delta. \tag{D.4}$$

This is because, if $\mathbf{D}$ satisfies $\mathcal{E}_{k-1}(\mathbf{D})$ then (D.4) comes from (D.3); or if $\mathbf{D}$ does not satisfy $\mathcal{E}_{k-1}(\mathbf{D})$ then the left hand side of (D.4) is zero (see Remark D.1) but the right hand side is *non-negative*.

Taking expectation with respect to the randomness of $\mathbf{D}$ in (D.4), and using Lemma D.3 which upper bounds $\mathbb{E}_\mathbf{D}[\mathbf{P}_k\mathbf{P}_k^\top \bullet \mathbf{X}_k]$ by $\mathbb{E}_\mathbf{D}[\mathbf{P}_k\mathbf{P}_k^\top \bullet \mathbf{X}_k\mathbf{D}] = \mathbb{E}_\mathbf{U}[\mathbf{P}_k\mathbf{P}_k^\top \bullet \mathbf{X}_k\mathbf{U}]$ we get the desired inequality. (Note that $\mathbf{P}_k\mathbf{P}_k^\top\mathbf{X}_k\mathbf{U} = \mathbf{A}_k\mathbf{X}_k\mathbf{U} = \mathbf{A}_k\mathbf{X}_k^{1/2}\mathbf{U}\mathbf{X}_k^{1/2}$.) □

---

[18]Note when $\mathbf{D}$ satisfies $\mathcal{E}_{k-1}(\mathbf{D})$ we have $c_k\mathbf{I} - \eta\mathbf{\Sigma}_{k-1} \succeq \frac{1}{e}\mathbf{I}$ according to Proposition 6.4. This implies, as long as $\eta \leq e^{-1}$, it satisfies $c_k\mathbf{I} - \eta\mathbf{\Sigma}_k \succ 0$ so $\mathbf{Tr}\Big((c_k\mathbf{I} - \eta\mathbf{\Sigma}_k)^{-(q-1)}\mathbf{U}\Big) > 0$.

## D.1 Missing Auxiliary Lemmas

In this subsection we prove the following two auxiliary lemmas. The first one shall be used to bound the higher-order terms in Lemma 5.1.

> **Lemma D.2.** *For every $q \geq 2$ and every $\eta \in \left[0, \frac{1}{4e(q-1)}\right]$, event $\mathcal{E}_{k-1}(\mathbf{D})$ implies that*
> $$\left|\mathbf{Tr}\left(\left(\mathbf{X}_k^{1/q} + \eta \mathbf{X}_k^{1/q}\mathbf{P}_k(\mathbf{I} - \eta \mathbf{P}_k^\top \mathbf{X}_k^{1/q}\mathbf{P}_k)^{-1}\mathbf{P}_k^\top \mathbf{X}_k^{1/q}\right)^{q-1}\mathbf{D}\right) - \mathbf{Tr}\left(\mathbf{X}_k^{\frac{q-1}{q}}\mathbf{D}\right)\right|$$
> $$\leq 4\eta(q-1)\|\mathbf{D}\|_2 \mathbf{Tr}(\mathbf{X}_k \mathbf{P}_k \mathbf{P}_k^\top) \ .$$

The second one upper bounds the expectation of the right hand side of Lemma D.2. We highlight that the proof of Lemma D.3 is the only place in this paper that we have assumed $k(\mathbf{U}) = 3$.

> **Lemma D.3.** *We have $\mathbb{E}_\mathbf{D}[\mathbf{Tr}(\mathbf{P}_k \mathbf{P}_k^\top \mathbf{X}_k)] \leq 9 \cdot \mathbb{E}_\mathbf{D}[\mathbf{Tr}(\mathbf{P}_k \mathbf{P}_k^\top \mathbf{X}_k \mathbf{D})]$.*

Note that we can assume without loss of generality that $\boldsymbol{\Sigma}_{k-1}$, $\mathbf{X}_k$ and $\mathbf{D}$ are all diagonal matrices, which has been argued in the proof of Lemma 6.5. Therefore, all the proofs in this subsection will be given under this assumption.

To prove Lemma D.2 we need the following lemma:

**Lemma D.4** (Monotonicity of Diagonal entries). *Let $\mathbf{A}, \mathbf{D} \in \mathbb{R}^{d \times d}$ be two diagonal positive definite matrices,[19] let $\mathbf{B} \in \mathbb{R}^{d \times d}$ be PSD, then for every $q \in \mathbb{N}^*$ such that $q\|\mathbf{A}^{-1/2}\mathbf{B}\mathbf{A}^{-1/2}\|_2 < 1$:*
$$0 \leq \mathbf{Tr}((\mathbf{A}+\mathbf{B})^q \mathbf{D}) - \mathbf{Tr}(\mathbf{A}^q \mathbf{D}) \leq \frac{\|\mathbf{D}\|_2}{1 - q\|\mathbf{A}^{-1/2}\mathbf{B}\mathbf{A}^{-1/2}\|_2} \mathbf{Tr}\left(\mathbf{A}^{q-1}\mathbf{B}\right) \ .$$

*Proof of Lemma D.4.* For every $i \in [D]$, let $\mathbf{P}$ be a matrix with all zero entries except $\mathbf{P}_{i,i} = 1$. Then we have:
$$\begin{aligned}
[(\mathbf{A}+\mathbf{B})^q]_{i,i} &= \mathbf{Tr}(\mathbf{P}^q(\mathbf{A}+\mathbf{B})^q \mathbf{P}^q) \geq \mathbf{Tr}\left((\mathbf{P}(\mathbf{A}+\mathbf{B})\mathbf{P})^q\right) \\
&= ([\mathbf{A}+\mathbf{B}]_{i,i})^q \geq [\mathbf{A}]_{i,i}^q = [\mathbf{A}^q]_{i,i} \ .
\end{aligned}$$

Where the first inequality is due to the Lieb-Thirring inequality, and the last equality is because $\mathbf{A}$ is diagonal. Since $\mathbf{D}$ is a diagonal PSD matrix, we can conclude that[20]
$$\mathbf{Tr}((\mathbf{A}+\mathbf{B})^q \mathbf{D}) - \mathbf{Tr}(\mathbf{A}^q \mathbf{D}) = \sum_{i=1}^d [\mathbf{D}]_{i,i}\left([(\mathbf{A}+\mathbf{B})^q - \mathbf{A}^q]_{i,i}\right) \geq 0 \ .$$

and
$$\mathbf{Tr}((\mathbf{A}+\mathbf{B})^q \mathbf{D}) - \mathbf{Tr}(\mathbf{A}^q \mathbf{D}) \leq \max_{i \in [d]}[\mathbf{D}]_{i,i} \sum_{i=1}^d [(\mathbf{A}+\mathbf{B})^q - \mathbf{A}^q]_{i,i} = \|\mathbf{D}\|_2 \mathbf{Tr}((\mathbf{A}+\mathbf{B})^q - \mathbf{A}^q) \ . \quad \text{(D.5)}$$

We focus on the term $(\mathbf{A}+\mathbf{B})^q$. We can re-write it as $(\mathbf{A}+\mathbf{B})^q = \left(\mathbf{A}^{1/2}(\mathbf{I} + \mathbf{A}^{-1/2}\mathbf{B}\mathbf{A}^{-1/2})\mathbf{A}^{1/2}\right)^q$. Then by Lieb-Thirring again, we have:
$$\begin{aligned}
\mathbf{Tr}((\mathbf{A}+\mathbf{B})^q) &\leq \mathbf{Tr}\left(\mathbf{A}^{q/2}\left(\mathbf{I} + \mathbf{A}^{-1/2}\mathbf{B}\mathbf{A}^{-1/2}\right)^q \mathbf{A}^{q/2}\right) \\
&\leq \mathbf{Tr}\left(\mathbf{A}^{q/2}\left(\mathbf{I} + \frac{1}{1 - q\|\mathbf{A}^{-1/2}\mathbf{B}\mathbf{A}^{-1/2}\|_2}\mathbf{A}^{-1/2}\mathbf{B}\mathbf{A}^{-1/2}\right)\mathbf{A}^{q/2}\right) \\
&\leq \mathbf{Tr}(\mathbf{A}^q) + \frac{q}{1 - q\|\mathbf{A}^{-1/2}\mathbf{B}\mathbf{A}^{-1/2}\|_2} \mathbf{Tr}\left(\mathbf{A}^{q-1}\mathbf{B}\right) \ . \quad \text{(D.6)}
\end{aligned}$$

---
[19] In fact, we have only required them to be simultaneously diagonalizable.
[20] The authors would like to thank Elliott Lieb who has helped us obtain the inequality of the next line.

Where the second inequality uses $(\mathbf{I} + \mathbf{X})^q \preceq \mathbf{I} + \frac{q}{1-q\|\mathbf{X}\|_2}\mathbf{X}$ for every PSD matrix $\mathbf{X}$ with $q\|\mathbf{X}\|_2 < 1$. Putting together (D.5) and (D.6), we obtain:

$$\mathbf{Tr}((\mathbf{A}+\mathbf{B})^q\mathbf{D}) - \mathbf{Tr}(\mathbf{A}^q\mathbf{D}) \leq \frac{q\|\mathbf{D}\|_2}{1-q\|\mathbf{A}^{-1/2}\mathbf{B}\mathbf{A}^{-1/2}\|_2}\mathbf{Tr}\left(\mathbf{A}^{q-1}\mathbf{B}\right) \ . \qquad \square$$

*Proof of Lemma D.2.* Under event $\mathcal{E}_{k-1}(\mathbf{D})$, we know $\mathbf{I}-\eta\mathbf{P}_k^\top\mathbf{X}_k^{1/q}\mathbf{P}_k \succeq (1-e\eta)\mathbf{I}$ (see Proposition 6.4) and thus

$$0 \preceq \eta\mathbf{X}_k^{1/2q}\mathbf{P}_k(\mathbf{I} - \eta\mathbf{P}_k^\top\mathbf{X}_k^{1/2q}\mathbf{P}_k)^{-1}\mathbf{P}_k^\top\mathbf{X}_k^{1/q} \preceq \frac{e\eta}{1-e\eta}\mathbf{I} \ .$$

We now apply Lemma D.4 with $\mathbf{A} = \mathbf{X}_k^{1/q}$, $\mathbf{B} = \eta\mathbf{X}_k^{1/q}\mathbf{P}_k(\mathbf{I}-\eta\mathbf{P}_k^\top\mathbf{X}_k^{1/q}\mathbf{P}_k)^{-1}\mathbf{P}_k^\top\mathbf{X}_k^{1/q}$, and $q = q-1$. We can do so because $\mathbf{A}$ and $\mathbf{D}$ are both diagonal and $\frac{(q-1)e\eta}{1-e\eta} < 1$ under our assumption of $\eta$. The conclusion of Lemma D.4 tells us that:

$$\left|\mathbf{Tr}\left(\left(\mathbf{X}_k^{1/q} + \eta\mathbf{X}_k^{1/q}\mathbf{P}_k(\mathbf{I}-\eta\mathbf{P}_k^\top\mathbf{X}_k^{1/q}\mathbf{P}_k)^{-1}\mathbf{P}_k^\top\mathbf{X}_k^{1/q}\right)^{q-1}\mathbf{D}\right) - \mathbf{Tr}\left(\mathbf{X}_k^{\frac{q-1}{q}}\mathbf{D}\right)\right|$$
$$\leq \frac{q-1}{1 - \frac{(q-1)e\eta}{1-e\eta}}\|\mathbf{D}\|_2 \mathbf{Tr}(\mathbf{A}^{q-2}\mathbf{B}) \leq \left(2(q-1)\|\mathbf{D}\|_2\right)\left(\frac{\eta}{1-e\eta}\mathbf{Tr}(\mathbf{X}_k\mathbf{P}_k\mathbf{P}_k^\top)\right)$$
$$\leq 4\eta(q-1)\|\mathbf{D}\|_2 \mathbf{Tr}(\mathbf{X}_k\mathbf{P}_k\mathbf{P}_k^\top) \ .$$

Above, the second and third inequalities have respectively used $\frac{(q-1)e\eta}{1-e\eta} < \frac{1}{2}$ and $\frac{1}{1-e\eta} \leq 2$, which are both true by our assumption on $\eta$. $\qquad \square$

*Proof of Lemma D.3.* Let $\lambda_1 \geq \cdots \geq \lambda_d \geq 0$ be the eigenvalues of $\mathbf{\Sigma}_{k-1}$ and $\nu_1, \ldots, \nu_d$ be the corresponding eigenvectors. Let $D_1, \cdots, D_d$ be the diagonals of $\mathbf{D}$. Recall that $\mathbf{\Sigma}_{k-1}, \mathbf{X}_k, \mathbf{D}$ are all diagonal matrices. Define function $f \colon \mathbb{R}^d \to \mathbb{R}$

$$f(r_1, \cdots, r_d) \stackrel{\text{def}}{=} \sum_{i=1}^d \frac{[\mathbf{P}_k\mathbf{P}_k^\top]_{i,i} \cdot r_i}{(c_k - \lambda_i)^q} \quad \text{(recall that } c_k \text{ depends on } (D_1, \ldots, D_d))$$

We shall prove that for some $\gamma \in (0,1)$ that shall be chosen later, it satisfies for every $i \in [d]$,

$$\mathbb{E}[f(\gamma, \cdots, \gamma, D_i, \cdots, D_d)] \geq \mathbb{E}[f(\gamma, \cdots, \gamma, D_{i+1}, \cdots, D_d)]$$

where recall that both expectations are only over the randomness of $D_1, \ldots, D_d$. Let $D_{-i} \stackrel{\text{def}}{=} (D_1, \ldots, D_i, D_{i+2}, \cdots, D_d)$. Then, it is sufficient to prove that for every fixed possibility of $D_{-i}$, the following inequality holds:

$$\mathbb{E}_{D_i}[f(\gamma, \cdots, \gamma, D_i, \cdots, D_d) \mid D_{-i}] \geq \mathbb{E}_{D_i}[f(\gamma, \cdots, \gamma, D_{i+1}, \cdots, D_d) \mid D_{-i}] \ .$$

Therefore, in the remaining proofs, we shall consider $D_i$ as the only random variable, and thus $c_k$ only depends on $D_i$. For a fixed value $s \geq 1$ that we shall choose later, we can let $c$ be the (unique) value of $c_k$ when $D_i = s\gamma$.

Letting $g(x) \stackrel{\text{def}}{=} \frac{x}{(c_k-\lambda_i)^q}$, we make three quick observations:

1. $g(\gamma) = \frac{\gamma}{(c_k-\lambda_i)^q}$ is a monotone decreasing function of $D_i$.

   This is so because $c_k$ is a monotone increasing function of $D_i$.

2. $g(D_i) = \frac{D_i}{(c_k-\lambda_i)^q}$ is a monotone decreasing function of $D_i$.

   This is because $g(D_i) = 1 - \sum_{j \neq i} \frac{D_j}{(c_k-\lambda_j)^q} = 1$ but $c_k$ is a monotone increasing function of $D_i$.

3. When $D_i \leq s\gamma$, we have $g(\gamma) \leq \frac{s\gamma}{D_i} \frac{\gamma}{(c-\lambda_i)^q}$.

   This is because $g(\gamma) = \frac{\gamma}{D_i}\left(1 - \sum_{j \neq i} \frac{D_j}{(c_k - \lambda_j)^q}\right) \leq \frac{\gamma}{D_i}\left(1 - \sum_{j \neq i} \frac{D_j}{(c-\lambda_j)^q}\right) = \frac{\gamma}{D_i} \frac{s\gamma}{(c-\lambda_j)^q}$, where the first inequality is because $c_k \leq c$ when $D_i \leq s\gamma$ (by the monotone increasing of $c_k$ with respect to $D_i$), and the second equality is according to the definition of $c$.

Combining the above three observations, we have:

$$\mathbb{E}[g(D_i)] \geq \mathbf{Pr}[D_i \geq s\gamma]\,\mathbb{E}[g(D_i) \mid D_i \geq s\gamma] \geq \mathbf{Pr}[D_i \geq s\gamma]\frac{s\gamma}{(c-\lambda_i)^q}$$

$$\mathbb{E}[g(\gamma)] \leq \mathbf{Pr}[D_i \geq s\gamma]\,\mathbb{E}[g(\gamma) \mid D_i \geq s\gamma] + \mathbb{E}\left[\frac{1}{D_i}\right]\frac{s\gamma^2}{(c-\lambda_i)^q}$$

$$\leq \frac{\gamma}{(c-\lambda_i)^q} + \mathbb{E}\left[\frac{1}{D_i}\right]\frac{s\gamma^2}{(c-\lambda_i)^q} \leq \frac{s\gamma}{(c-\lambda_i)^q}\left(\frac{1}{s} + \mathbb{E}\left[\frac{\gamma}{D_i}\right]\right) \ .$$

Recall that each $D_i = \frac{1}{3}(\langle \nu_i, u_1\rangle^2 + \langle \nu_i, u_2\rangle^2 + \langle \nu_i, u_3\rangle^2)$ where $u_1, u_2, u_3$ are three normal Gaussian random vectors. Therefore, each $3D_i$ has a chi-square distribution of degree 3, which implies $\mathbb{E}[\frac{1}{D_i}] = 3$ and $\mathbf{Pr}[D_i \geq \frac{1}{3}] > \frac{2}{3}$. In sum, if we take $\gamma = \frac{1}{9}$ and $s = 3$, we have:

$$\mathbb{E}_{D_i}[g(D_i)] \geq \mathbb{E}_{D_i}[g(\gamma)] \ .$$

Finally, this implies

$$\mathbb{E}_{D_i}[f(\gamma, \cdots, \gamma, D_i, \cdots, D_d) - f(\gamma, \cdots, \gamma, D_{i+1}, \cdots, D_d) \mid D_{-i}] = [\mathbf{P}_k \mathbf{P}_k^\top]_{i,i}\, \mathbb{E}_{D_i}[g(D_i) - g(\gamma) \mid D_{-i}] \geq 0 \ .$$

so we have

$$\mathbb{E}_{\mathbf{D}}[f(\gamma, \cdots, \gamma, D_i, \cdots, D_d)] \geq \mathbb{E}_{\mathbf{D}}[f(\gamma, \cdots, \gamma, D_{i+1}, \cdots, D_d)] \ .$$

In particular,

$$\mathbb{E}[\mathbf{Tr}(\mathbf{P}_k \mathbf{P}_k^\top \mathbf{X}_k \mathbf{D})] = \mathbb{E}[f(D_1, \cdots, D_d)] \geq \mathbb{E}[f(\gamma, \cdots, \gamma)] = \gamma\, \mathbb{E}[\mathbf{Tr}(\mathbf{P}_k \mathbf{P}_k^\top \mathbf{X}_k)] \ . \qquad \square$$

# E  Proof for Section 6.3

**Lemma 6.6.** *For all $q \geq 2$ and $\eta > 0$,*

$$\mathbb{E}\left[\mathbf{Tr}\left((c_{k+1}\mathbf{I} - \eta\mathbf{\Sigma}_k)^{-(q-1)}\mathbf{U}\right) \cdot \mathbb{1}_{\mathcal{E}_{<(k+1)}(\mathbf{U})}\right] - \mathbb{E}\left[\mathbf{Tr}\left((c_k\mathbf{I} - \eta\mathbf{\Sigma}_k)^{-(q-1)}\mathbf{U}\right) \cdot \mathbb{1}_{\mathcal{E}_{<k}(\mathbf{U})}\right]$$
$$\leq -(q-1)(\mathbb{E}[c_{k+1}] - \mathbb{E}[c_k])$$

*Proof.* Recall that $c_{k+1} \geq c_k$ because all matrices $\mathbf{A}_k$ are PSD. Denoting by $\nu_1, \ldots, \nu_d$ the eigenvectors of $\mathbf{\Sigma}_k$ with non-increasing eigenvalues $\lambda_1 \geq \cdots \geq \lambda_d$,[21] we have for every $\mathbf{U}$,

$$\mathbf{Tr}\left((c_{k+1}\mathbf{I} - \eta\mathbf{\Sigma}_k)^{-(q-1)}\mathbf{U}\right) - \mathbf{Tr}\left((c_k\mathbf{I} - \eta\mathbf{\Sigma}_k)^{-(q-1)}\mathbf{U}\right)$$
$$= \sum_{i=1}^d \frac{\nu_i^\top \mathbf{U} \nu_i}{(c_{k+1} - \eta\lambda_i)^{q-1}} - \sum_{i=1}^d \frac{\nu_i^\top \mathbf{U} \nu_i}{(c_k - \eta\lambda_i)^{q-1}} \overset{①}{\leq} -(q-1)(c_{k+1} - c_k) \cdot \sum_{i=1}^d \frac{\nu_i^\top \mathbf{U} \nu_i}{(c_{k+1} - \eta\lambda_i)^q}$$
$$= -(q-1)(c_{k+1} - c_k) \cdot \mathbf{Tr}\left((c_{k+1}\mathbf{I} - \eta\mathbf{\Sigma}_k)^{-q}\mathbf{U}\right) = -(q-1)(c_{k+1} - c_k)\mathbf{Tr}(\mathbf{X}_{k+1}\mathbf{U})$$
$$= -(q-1)(c_{k+1} - c_k) \ . \tag{E.1}$$

---
[21]This is different from the proof of Lemma 6.5 where we defined them to be eigenvectors of $\mathbf{\Sigma}_{k-1}$.

Above, inequality ① is derived from inequality $\frac{1}{(c+x)^{q-1}} - \frac{1}{x^{q-1}} \leq -\frac{(q-1)c}{(c+x)^q}$ (for every $c \geq 0$, $x > 0$) which follows from the convexity of function $f(x) = \frac{1}{x^{q-1}}$.

Next, we observe that for every $\mathbf{U}$ that does *not* satisfy $\mathcal{E}_{<k}(\mathbf{U})$, the very right hand side of (E.1) is still non-negative. Therefore, we conclude that for all $\mathbf{U}$,

$$\mathbf{Tr}\Big((c_{k+1}\mathbf{I} - \eta\boldsymbol{\Sigma}_k)^{-(q-1)}\mathbf{U}\Big) \cdot \mathbb{1}_{\mathcal{E}_{<k}(\mathbf{U})} - \mathbf{Tr}\Big((c_k\mathbf{I} - \eta\boldsymbol{\Sigma}_k)^{-(q-1)}\mathbf{U}\Big) \cdot \mathbb{1}_{\mathcal{E}_{<k}(\mathbf{U})} \leq -(q-1)(c_{k+1} - c_k) \ .$$

Finally, since $\mathbb{1}_{\mathcal{E}_{<(k+1)}(\mathbf{U})} \leq \mathbb{1}_{\mathcal{E}_{<k}(\mathbf{U})}$ and $\mathbf{Tr}\Big((c_k\mathbf{I} - \eta\boldsymbol{\Sigma}_k)^{-(q-1)}\mathbf{U}\Big) \geq 0$, we have

$$\mathbf{Tr}\Big((c_{k+1}\mathbf{I} - \eta\boldsymbol{\Sigma}_k)^{-(q-1)}\mathbf{U}\Big) \cdot \mathbb{1}_{\mathcal{E}_{<(k+1)}(\mathbf{U})} - \mathbf{Tr}\Big((c_k\mathbf{I} - \eta\boldsymbol{\Sigma}_k)^{-(q-1)}\mathbf{U}\Big) \cdot \mathbb{1}_{\mathcal{E}_{<k}(\mathbf{U})} \leq -(q-1)(c_{k+1} - c_k)$$

and taking expectation we finish the proof of Lemma 6.6. $\square$

## F  Proof of Theorem 1: Oblivious Online Eigenvector

**Theorem 1.** *In the online eigenvector problem with an **oblivious** adversary, there exists absolute constant $C > 1$ such that if $q \geq 3\log(2dT)$ and $\eta \in \big[0, \frac{1}{11q^3}\big]$, then $\mathtt{FTCL}^{\mathsf{obl}}(T, q, \eta)$ satisfies*

$$\sum_{k=1}^{T} \mathbb{E}\big[w_k^\top \mathbf{A}_k w_k\big] = \sum_{k=1}^{T} \mathbb{E}\big[\mathbf{A}_k \bullet \mathbf{X}_k^{1/2}\mathbf{U}\mathbf{X}_k^{1/2}\big] \geq \big(1 - C \cdot \eta q^5 \log(dT)\big)\lambda_{\max}(\boldsymbol{\Sigma}_T) - \frac{4}{\eta} \ .$$

*Proof of Theorem 1.* Combining Lemma 6.5 and Lemma 6.6, we have

$$\mathbb{E}\Big[\mathbf{Tr}\big(\mathbf{X}_{k+1}^{1-1/q}\mathbf{U}\big) \cdot \mathbb{1}_{\mathcal{E}_{<k+1}(\mathbf{U})}\Big] - \mathbb{E}\Big[\mathbf{Tr}\big(\mathbf{X}_k^{1-1/q}\mathbf{U}\big) \cdot \mathbb{1}_{\mathcal{E}_{<k}(\mathbf{U})}\Big]$$
$$\leq -(q-1)(\mathbb{E}[c_{k+1}] - \mathbb{E}[c_k]) + (q-1)\eta(1 + O(\eta q^5 \log(d/\delta))) \cdot \mathbb{E}\big[\mathbf{A}_k \bullet \mathbf{X}_k^{1/2}\mathbf{U}\mathbf{X}_k^{1/2}\big] + (\eta T + e)T\delta \ .$$

Telescoping it for all $k = 1, \ldots, T$, we have

$$\mathbb{E}\Big[\mathbf{Tr}\big(\mathbf{X}_{T+1}^{1-1/q}\mathbf{U}\big) \cdot \mathbb{1}_{\mathcal{E}_{<T+1}(\mathbf{U})}\Big] - \mathbb{E}\Big[\mathbf{Tr}\big(\mathbf{X}_1^{1-1/q}\mathbf{U}\big) \cdot \mathbb{1}_{\mathcal{E}_{<1}(\mathbf{U})}\Big] \tag{F.1}$$
$$\leq -(q-1)(\mathbb{E}[c_{T+1}] - \mathbb{E}[c_1]) + (q-1)\eta(1 + O(\eta q^5 \log(d/\delta))) \cdot \mathbb{E}\Big[\sum_{k=1}^{T} \mathbf{A}_k \bullet \mathbf{X}_k^{1/2}\mathbf{U}\mathbf{X}_k^{1/2}\Big] + (\eta T + e)T^2\delta \ .$$

We make four quick observations:

- Regardless of the randomness of $\mathbf{U}$, we have $\mathbf{Tr}\big(\mathbf{X}_{T+1}^{1-1/q}\mathbf{U}\big) \cdot \mathbb{1}_{\mathcal{E}_{<T+1}(\mathbf{U})} \geq 0$.
- Regardless of the randomness of $\mathbf{U}$, we have $c_{T+1} \geq \eta\lambda_{\max}(\boldsymbol{\Sigma}_T)$.
- We have $\mathbb{E}[c_1] \leq e$. To derive that, we use $\frac{1}{c_1^q}\mathbf{Tr}\mathbf{U} = \mathbf{Tr}(\mathbf{X}_1\mathbf{U}) = 1$ which implies $c_1 = (\mathbf{Tr}\mathbf{U})^{1/q}$. Notice that $\mathbf{Tr}\mathbf{U} = \frac{1}{3}\sum_{i \in [d], j \in [3]}(u_{j,i})^2$ so $3\mathbf{Tr}\mathbf{U}$ is distributed according to chi-squared distribution $\chi^2(3d)$ whose PDF is $p(x) = \frac{2^{-\frac{3d}{2}}e^{-\frac{x}{2}}x^{\frac{3d}{2}-1}}{\Gamma(3d/2)}$. We thus have

$$\mathbb{E}[c_1] = \int_0^\infty x^{1/q}p(x)dx = \frac{2^{1/q}\Gamma\big(\frac{3d}{2} + \frac{1}{q}\big)}{\Gamma\big(\frac{3d}{2}\big)} \leq 2^{1/q} \cdot \big(\frac{3d}{2}\big)^{1/q} = (3d)^{1/q} \leq e \ .$$

Above, the first inequality uses $\frac{\Gamma(x+a)}{\Gamma(x)} \leq x^a$ for $a \in (0, 1)$ and $x > 0$ (cf. Wendell [39]), and the second inequality uses our assumption on $q$.

- $\mathbb{E}\Big[\mathbf{Tr}\big(\mathbf{X}_1^{1-1/q}\mathbf{U}\big) \cdot \mathbb{1}_{\mathcal{E}_{<1}(\mathbf{U})}\Big] \leq e$. This is because $\mathbf{Tr}(\mathbf{X}_1^{1-1/q}\mathbf{U}) = \frac{1}{c_1^{q-1}}\mathbf{Tr}\mathbf{U} = c_1$ and $\mathbb{E}[c_1] \leq e$.

Substituting the four observations above into the telescoping sum (F.1), we have

$$(q-1)\eta\lambda_{\max}(\mathbf{\Sigma}_T) \leq e + (q-1)e + (q-1)\eta(1 + O(\eta q^5 \log(d/\delta))) \cdot \mathbb{E}\Big[\sum_{k=1}^{T} \mathbf{A}_k \bullet \mathbf{X}_k^{1/2}\mathbf{U}\mathbf{X}_k^{1/2}\Big] + (\eta T + e)T^2\delta \ .$$

Using the inequality $(\eta T + e)T^2\delta \leq (1+e)T^3\delta$, we conclude that if we choose $\delta = \frac{1}{1+e}T^{-3}$, then

$$(q-1)\eta\lambda_{\max}(\mathbf{\Sigma}_T) \leq (q-1)\eta \left(1 + O(\eta q^5 \log(dT))\right) \cdot \mathbb{E}\Big[\sum_{k=1}^{T} \mathbf{A}_k \bullet \mathbf{X}_k^{1/2}\mathbf{U}\mathbf{X}_k^{1/2}\Big] + 4(q-1) \ .$$

Dividing both sides by $(q-1)\eta$, and recalling that $\mathbb{E}[w_k w_k^\top] = \mathbf{X}_k^{1/2}\mathbf{U}\mathbf{X}_k^{1/2}$, we arrive at the desired inequality. $\square$

## G  Proof of Theorem 2: Adversarial Online Eigenvector

**Theorem 2.** *In the online eigenvector problem with an **adversarial** adversary, there exists constant $C > 1$ such that for every $p \in (0,1)$, $q \geq 3\log(2dT)$ and $\eta \in \left[0, \frac{1}{11q^3}\right]$, our* $\texttt{FTCL}^{\mathsf{adv}}(T, q, \eta)$ *satisfies*

$$w.p. \geq 1-p: \quad \sum_{k=1}^{T} w_k^\top \mathbf{A}_k w_k \geq \Big(1 - C \cdot \eta\big(q^5\log(dT) + \log(1/p)\big)\Big)\lambda_{\max}(\mathbf{\Sigma}_T) - \frac{5}{\eta} \ .$$

*Proof of Theorem 2.* Before beginning our proof, let us emphasize that in this adversarial setting,

- $\mathbf{A}_k$ and $\mathbf{\Sigma}_k$ can depend on the randomness of $\mathbf{U}_1, \ldots, \mathbf{U}_{k-1}$.
- $\mathbf{X}_k$ and $c_k$ depend on the randomness of $\mathbf{U}_k$ and $\mathbf{\Sigma}_{k-1}$ (and thus also on $\mathbf{U}_1, \ldots, \mathbf{U}_{k-2}$).

Consider (for analysis purpose only) another random matrix $\widetilde{\mathbf{U}}$ drawn from distribution $\mathcal{D}$, independent of the randomness of $\mathbf{U}_1, \ldots, \mathbf{U}_T$. Define $\widetilde{c}_k$ to be the unique constant satisfying $\widetilde{c}_k \mathbf{I} - \eta \mathbf{\Sigma}_{k-1} \succ 0$ and $\mathbf{Tr}((\widetilde{c}_k \mathbf{I} - \eta \mathbf{\Sigma}_{k-1})^{-q}\mathbf{U}) = 1$, and define $\widetilde{\mathbf{X}}_k = (\widetilde{c}_k \mathbf{I} - \eta \mathbf{\Sigma}_{k-1})^{-q}$.

Now, if we fix the randomness of $\mathbf{U}_1, \ldots, \mathbf{U}_{k-1}$, the matrices $\mathbf{\Sigma}_{k-1}$ and $\mathbf{A}_k$ become fixed. The fact that $\mathbf{U}_k$ and $\widetilde{\mathbf{U}}$ are both drawn from the same distribution $\mathcal{D}$ (and the fact that $\mathbf{X}_k$ and $\widetilde{\mathbf{X}}_k$ are computed from $\mathbf{U}_k$ and $\widetilde{\mathbf{U}}$ in the same way) implies

$$\mathbb{E}_{\mathbf{U}_k}\Big[\mathbf{A}_k \bullet \mathbf{X}_k^{1/2}\mathbf{U}_k\mathbf{X}_k^{1/2}\Big|\mathbf{U}_1,\ldots,\mathbf{U}_{k-1}\Big] = \mathbb{E}_{\widetilde{\mathbf{U}}}\Big[\mathbf{A}_k \bullet \widetilde{\mathbf{X}}_k^{1/2}\widetilde{\mathbf{U}}\widetilde{\mathbf{X}}_k^{1/2}\Big|\mathbf{U}_1,\ldots,\mathbf{U}_{k-1}\Big] \quad (\text{G.1})$$

Now, consider random variables $Z_k = w_k^\top \mathbf{A}_k w_k$. We have that $Z_k$ is $\mathcal{F}_k$-measurable for $\mathcal{F}_k$ generated by $\mathbf{U}_1, \ldots, \mathbf{U}_k, w_1, \ldots, w_k$. According to the martingale concentration Lemma G.1, we have

$$\mathbf{Pr}\left[\sum_{k=1}^{T} Z_k \leq (1-\mu)\sum_{k=1}^{T} \mathbb{E}[Z_k \mid \mathcal{F}_{k-1}] - \frac{\log \frac{1}{p}}{\mu}\right] \leq p \ .$$

At the same time, we have

$$\mathbb{E}[Z_k \mid \mathcal{F}_{k-1}] = \mathbb{E}_{w_k, \mathbf{U}_k}\Big[\mathbf{A}_k \bullet w_k w_k^\top \mid \mathbf{U}_1, \ldots, \mathbf{U}_{k-1}\Big] = \mathbb{E}_{\mathbf{U}_k}\Big[\mathbf{A}_k \bullet \mathbf{X}_k^{1/2}\mathbf{U}_k\mathbf{X}_k^{1/2} \mid \mathbf{U}_1, \ldots, \mathbf{U}_{k-1}\Big]$$

$$= \mathbb{E}_{\widetilde{\mathbf{U}}}\Big[\mathbf{A}_k \bullet \widetilde{\mathbf{X}}_k^{1/2}\widetilde{\mathbf{U}}\widetilde{\mathbf{X}}_k^{1/2} \mid \mathbf{U}_1, \ldots, \mathbf{U}_{k-1}\Big] \ ,$$

where the last inequality comes from (G.1). In sum, with probability at least $1 - p$ (over the randomness of $\mathbf{U}_1, \ldots, \mathbf{U}_T, w_1, \ldots, w_T$), we have

$$\sum_{k=1}^{T} w_k^\top \mathbf{A}_k w_k \geq (1-\mu) \mathop{\mathbb{E}}_{\widetilde{\mathbf{U}}} \Big[ \sum_{k=1}^{T} \mathbf{A}_k \bullet \widetilde{\mathbf{X}}_k^{1/2} \widetilde{\mathbf{U}} \widetilde{\mathbf{X}}_k^{1/2} \Big| \mathbf{U}_1, \ldots, \mathbf{U}_{T-1} \Big] - \frac{\log \frac{1}{p}}{\mu} \ .$$

Applying Theorem 1 we have (more specifically, fixing each possible sequence $\mathbf{U}_1, \ldots, \mathbf{U}_T$, we have a fixed sequence of $\mathbf{A}_1, \ldots, \mathbf{A}_T$ and can apply Theorem 1):

$$\sum_{k=1}^{T} w_k^\top \mathbf{A}_k w_k \geq (1-\mu)\big(1 - O(\eta q^5 \log(dT))\big) \lambda_{\max}(\mathbf{\Sigma}_T) - \frac{4}{\eta} - \frac{\log \frac{1}{p}}{\mu} \ .$$

Choosing $\mu = \eta \cdot \log(1/p)$, we finish the proof of Theorem 2. $\square$

## G.1 A Concentration Inequality for Martingales

We show the following (simple) martingale concentration lemma that we believe is classical but have not found anywhere else.

**Lemma G.1** (Concentration). *Let $\{Z_t\}_{t=1}^T$ be a random process with respect to a filter $\{0, \Omega\} = \mathcal{F}_0 \subset \mathcal{F}_1 \subset \cdots \subset \mathcal{F}_T$ and each $Z_t \in [0, 1]$ is $\mathcal{F}_t$-measurable. For every $p, \mu \in (0, 1)$,*

$$\mathbf{Pr}\left[ \sum_{t=1}^{T} Z_t \leq (1-\mu) \sum_{t=1}^{T} \mathbb{E}[Z_t \mid \mathcal{F}_{t-1}] - \frac{\log \frac{1}{p}}{\mu} \right] \leq p \ .$$

We emphasize here that $\mathbb{E}[Z_t \mid \mathcal{F}_{t-1}]$ is $\mathcal{F}_{t-1}$-measurable and thus not a constant.

*Proof of Lemma G.1.* Like in classical concentration proofs, we have

$$\mathbf{Pr}\left[ \sum_{t=1}^T Z_t \leq (1-\mu) \sum_{t=1}^T \mathbb{E}[Z_t \mid \mathcal{F}_{t-1}] - \frac{\log \frac{1}{p}}{\mu} \right]$$
$$= \mathbf{Pr}\left[ \sum_{t=1}^T \big((1-\mu) \mathbb{E}[Z_t \mid \mathcal{F}_{t-1}] - Z_t\big) \geq \frac{\log \frac{1}{p}}{\mu} \right]$$
$$= \mathbf{Pr}\left[ \exp\left\{ \mu \left( \sum_{t=1}^T \big((1-\mu) \mathbb{E}[Z_t \mid \mathcal{F}_{t-1}] - Z_t\big) \right) \right\} \geq \frac{1}{p} \right]$$
$$\leq p \, \mathbb{E}\left[ \exp\left\{ \mu \left( \sum_{t=1}^T \big((1-\mu) \mathbb{E}[Z_t \mid \mathcal{F}_{t-1}] - Z_t\big) \right) \right\} \right] \ . \tag{G.2}$$

Denote by $Y_t = \mu(1-\mu) \mathbb{E}[Z_t \mid \mathcal{F}_{t-1}] - \mu Z_t$, we know that each $Y_t \in [-1, 1]$ is $\mathcal{F}_t$-measurable.

$$\mathbb{E}\left[\exp\left\{\sum_{t=1}^T Y_t\right\}\right] = \mathbb{E}\left[\mathbb{E}\left[\exp\left\{\sum_{t=1}^T Y_t\right\} \big| \mathcal{F}_{T-1}\right]\right]$$
$$= \mathbb{E}\left[\exp\left\{\sum_{t=1}^{T-1} Y_t\right\} \mathbb{E}\left[e^{Y_T} \big| \mathcal{F}_{T-1}\right]\right]$$
$$\leq \mathbb{E}\left[\exp\left\{\sum_{t=1}^{T-1} Y_t\right\} \mathbb{E}\left[1 + Y_T + Y_T^2 \big| \mathcal{F}_{T-1}\right]\right] \ .$$

Now, we focus on the term $Y_T + Y_T^2$:

$$Y_T + Y_T^2 \leq \mu(1-\mu) \mathbb{E}[Z_T \mid \mathcal{F}_{T-1}] - \mu Z_T + \mu^2 (1-\mu)^2 \mathbb{E}[Z_T \mid \mathcal{F}_{T-1}]^2 + \mu^2 Z_T^2$$
$$\leq \mu(1-\mu) \mathbb{E}[Z_T \mid \mathcal{F}_{T-1}] - \mu Z_T + \mu^2 (\mathbb{E}[Z_T \mid \mathcal{F}_{T-1}] + \mu Z_T) \ .$$

(The first inequality has used $(a-b)^2 \leq a^2 + b^2$ when $a, b \geq 0$, and the second has used $Z_t \in [0, 1]$.)

Taking the conditional expectation, we obtain $\mathbb{E}[Y_T + Y_T^2 \mid \mathcal{F}_{T-1}] \leq 0$ and this implies

$$\mathbb{E}\left[\exp\left\{\sum_{t=1}^T Y_t\right\}\right] \leq \mathbb{E}\left[\exp\left\{\sum_{t=1}^{T-1} Y_t\right\}\right] \leq \cdots \leq e^0 = 1 \ .$$

Plugging this into (G.2) completes the proof of Lemma G.1. $\square$

# H  Proof of Theorem 3: Implementation Details

> **Theorem 3.** *If $q \geq 3\log(2dT/p)$, with probability at least $1-p$, for all $k \in [T]$, the $k$-th iteration of* FTCL[obl] *and* FTCL[adv] *runs in $O(d)$ plus the time to solve $\widetilde{O}(1)$ linear systems for matrices $c\mathbf{I} - \eta\boldsymbol{\Sigma}_{k-1}$. Here, $c > 0$ is some constant satisfying $c\mathbf{I} - \eta\boldsymbol{\Sigma}_{k-1} \succ \frac{1}{e}\mathbf{I}$.*

**Resolution to Issue (a).** We first point out that the final regret blows up by an additive value $\widetilde{\varepsilon}$ as long as the eigendecomposition $\sum_{j=1}^{3} p_j \cdot y_j y_j^\top$ is computed to satisfy[22]

$$\Big\|\sum_{j=1}^{3}\mathbf{X}_k^{1/2} u_j u_j^\top \mathbf{X}_k^{1/2} - \sum_{j=1}^{3} p_j \cdot y_j y_j^\top\Big\|_2 \leq \frac{\widetilde{\varepsilon}}{\mathsf{poly}(d,T)} \ .$$

Moreover, this can be done in time $O(d)$ as long as we can compute the three vectors $\{\mathbf{X}^{1/2} u_j\}_{j \in [3]}$ to an additive $\widetilde{\varepsilon}/\mathsf{poly}(d,T)$ error in Euclidean norm. This can be done by applying $(c_k\mathbf{I} - \eta\boldsymbol{\Sigma}_{k-1})^{-1}$ a number $q/2$ times to vector $u_j$, each again to an error $\widetilde{\varepsilon}/\mathsf{poly}(d,T)$. In sum, we can repeatedly apply Lemma H.1 and the final running time only logarithmically depends on $\widetilde{\varepsilon}/\mathsf{poly}(d,T)$.

**Resolution to Issue (c).** We choose $\delta = p/T$ and revisit the event $\mathcal{E}_k(\mathbf{U})$ defined in Def. 6.2. According to Lemma 6.3 and union bound, it satisfies with probability at least $1-p$, all the $T$ events $\mathcal{E}_0(\mathbf{U}_1),\ldots,\mathcal{E}_{T-1}(\mathbf{U}_T)$ are satisfied. If we apply Proposition 6.4, we immediately have that

$$q \geq 3\log(2dT/p) \implies \forall k \in [T]: \quad (\eta\lambda_{\max}(\boldsymbol{\Sigma}_{k-1}) + e)\mathbf{I} \succeq c_k\mathbf{I} \succeq c_k\mathbf{I} - \eta\boldsymbol{\Sigma}_{k-1} \succeq \frac{1}{e}\mathbf{I} \ . \tag{H.1}$$

This implies, throughout the algorithm, whenever we want to compute $(c_k\mathbf{I} - \eta\boldsymbol{\Sigma}_{k-1})^{-1}$, the matrix under inversion has a bounded condition number. We have the following lemma which relies on classical results from convex optimization:

**Lemma H.1.** *Given any $b \in \mathbb{R}^d$, the computation of $a \in \mathbb{R}^d$ satisfying $\|a - (c_k\mathbf{I} - \eta\boldsymbol{\Sigma}_{k-1})^{-1} b\|_2 \leq \varepsilon \|b\|_2$ can be done in running time*

- $\widetilde{O}\big(\sqrt{\eta\lambda_{\max}(\boldsymbol{\Sigma}_{k-1}) + 1} \cdot \mathsf{nnz}(\boldsymbol{\Sigma}_{k-1}) \cdot \log \varepsilon^{-1}\big)$ *if conjugate gradient or accelerated gradient descent is used;*
- $\widetilde{O}\big((\mathsf{nnz}(\boldsymbol{\Sigma}_{k-1}) + \sqrt{\eta k} \cdot \max_{i \in [k-1]} \{\mathsf{nnz}(\boldsymbol{\Sigma}_{k-1})^{3/4} \mathsf{nnz}(\mathbf{A}_i)^{1/4}\}) \log \varepsilon^{-1}\big)$ *if accelerated SVRG is used.*

*Proof.* This inverse operation is the same as minimizing a convex function $f(x) \stackrel{\text{def}}{=} \frac{1}{2} x^\top (c_k\mathbf{I} - \eta\boldsymbol{\Sigma}_{k-1}) x - b^\top x$. The condition number of Hessian matrix $\nabla^2 f(x)$ is at most $O(\eta\lambda_{\max}(\boldsymbol{\Sigma}_{k-1}) + 1)$ according to (H.1), so one can apply conjugate gradient [38] or Nesterov's accelerated gradient descent [29] to minimize this objective.

As for the SVRG type of result, one can write $f(x) = \frac{1}{k-1}\sum_{i=1}^{k-1} f_i(x)$ where $f_i(x) = x^\top (c_k\mathbf{I} - \eta(k-1)\mathbf{A}_i) x - b^\top x$. Each computation of $\nabla f(x)$ requires time $O(\mathsf{nnz}(\boldsymbol{\Sigma}_{k-1}))$ and that of $\nabla f_i(x)$ requires time $O(\mathsf{nnz}(\mathbf{A}_i))$. Since $\|\nabla^2 f_i(x)\|_2 \leq \eta k$ for each $i$, one can apply the SVRG method [10, 36] to minimize $f(x)$ which gives running time $\widetilde{O}\big(\mathsf{nnz}(\boldsymbol{\Sigma}_{k-1}) + (\eta k)^2 \max_{i \in [k-1]}\{\mathsf{nnz}(\mathbf{A}_i)\}\big)$. Then, using the Catalyst/APPA acceleration scheme [18, 28], the above running time can be improved to $\widetilde{O}\big(\mathsf{nnz}(\boldsymbol{\Sigma}_{k-1}) + \sqrt{\eta k} \cdot \max_{i \in [k-1]}\{\mathsf{nnz}(\boldsymbol{\Sigma}_{k-1})^{3/4} \mathsf{nnz}(\mathbf{A}_i)^{1/4}\}\big)$. □

---

[22]We refrain from doing this precisely here because because MMWU analysis is generally "robust against noise". The authors of [9] have shown that the potential $\mathbf{Tr}(\mathbf{X}_k^{1-1/q})$ is robust against noise and a completely analogous (but lengthy) proof of theirs applies to this paper.

**Resolution to Issue (b).** In each iteration, we need to compute some constant $c_k$ such that $\mathbf{Tr}(\mathbf{X}^{1/2}\mathbf{U}\mathbf{X}^{1/2}) = 1$. This can be done via a "binary search" procedure which was used widely for shift-and-invert based methods [19]:

1. Begin with $c = \eta k + e$ which is a safe upper bound on $c_k$ according to (H.1).

2. Repeatedly compute some value $\widetilde{\sigma}$ which is a 9/10 approximation of $\sigma \stackrel{\text{def}}{=} c - \eta\lambda_{\max}(\mathbf{\Sigma}_{k-1})$. (This requires $O(1)$ iterations of power method applied to $(c\mathbf{I} - \eta\mathbf{\Sigma}_{k-1})^{-1}$ [19].)

3. If $\widetilde{\sigma} \leq \frac{1}{e} \cdot \frac{9}{10}$ (which implies $\sigma \leq \frac{1}{e}$), we end the procedure; otherwise we update $c \leftarrow c - \widetilde{\sigma}/2$ and go to Step 2.

It is a simple exercise (with details given in [19]) to show that when the procedure ends, it satisfies $\frac{1}{2e} \leq c - \eta\mathbf{\Sigma}_{k-1} \leq \frac{1}{e}$ so $c$ is a lower bound on $c_k$. At this point, it suffices to perform a binary search between $[c, \eta k + e]$ to find $c_k$. Note that, according to resolution to issue (a), it suffices to compute $c_k$ to an additive error of $\widetilde{\varepsilon}/\mathsf{poly}(d, T)$.

In sum, the above binary search procedure requires only a logarithmic number of oracle calls to $(c\mathbf{I} - \eta\mathbf{\Sigma}_{k-1})^{-1}$, and each time we do so it satisfies $c \leq \eta k + e$ and $(\eta k + e)\mathbf{I} \succeq c\mathbf{I} - \eta\mathbf{\Sigma}_{k-1} \succeq \frac{1}{2e}\mathbf{I}$. For this reason, the same computational complexity in Lemma H.1 applies.

The three resolutions above, combined together, imply that the running time statements in Theorem 3 hold.

## I  Proof of Theorem 4: Stochastic Online Eigenvector

> **Theorem 4.** *There exists $C > 1$ such that, for every $p \in (0,1)$, if $\eta \in \left[0, \sqrt{p/(75T\lambda_{\max}(\mathbf{B}))}\right]$ in Oja's algorithm, we have with probability at least $1 - p$:*
> $$\sum_{k=1}^T w_k^\top \mathbf{A}_k w_k \geq (1 - 2\eta)T \cdot \lambda_{\max}(\mathbf{B}) - C \cdot \frac{\log(d/p)}{\eta} \ .$$

*Proof of Theorem 4.* Let $\lambda$ and $\nu$ be the largest eigenvalue and the corresponding normalized eigenvector of $\mathbf{B}$, and define

$$\Phi_k^{\mathbf{M}} \stackrel{\text{def}}{=} (\mathbf{I} + \eta\mathbf{A}_k)\cdots(\mathbf{I} + \eta\mathbf{A}_1)\mathbf{M}(\mathbf{I} + \eta\mathbf{A}_1)\cdots(\mathbf{I} + \eta\mathbf{A}_k)$$
$$\Psi_k \stackrel{\text{def}}{=} (\mathbf{I} + \eta\mathbf{A}_k)\cdots(\mathbf{I} + \eta\mathbf{A}_T)\nu\nu^\top(\mathbf{I} + \eta\mathbf{A}_T)\cdots(\mathbf{I} + \eta\mathbf{A}_k) \ .$$

We first make three calculations:

$$\mathbf{Tr}(\Phi_T^{uu^\top}) = \mathbf{Tr}\big((\mathbf{I} + \eta\mathbf{A}_T)\Phi_{T-1}^{uu^\top}(\mathbf{I} + \eta\mathbf{A}_T)\big) = \mathbf{Tr}(\Phi_{T-1}^{uu^\top}) + 2\eta\mathbf{Tr}\big(\mathbf{A}_T\Phi_{T-1}^{uu^\top}\big) + \eta^2\mathbf{Tr}\big(\mathbf{A}_T\Phi_{T-1}^{uu^\top}\mathbf{A}_T\big)$$
$$\overset{\text{①}}{\leq} \mathbf{Tr}(\Phi_{T-1}^{uu^\top}) \cdot (1 + (2\eta + \eta^2)\mathbf{Tr}(\mathbf{A}_T w_T w_T^\top)) \overset{\text{②}}{\leq} \mathbf{Tr}(\Phi_{T-1}^{uu^\top}) \cdot e^{(2\eta+\eta^2)w_T^\top \mathbf{A}_T w_T}$$
$$\leq \cdots \leq \mathbf{Tr}(\Phi_0^{uu^\top}) \cdot e^{(2\eta+\eta^2)\sum_{k=1}^T w_k^\top \mathbf{A}_k w_k} = \|u\|_2^2 \cdot e^{(2\eta+\eta^2)\sum_{k=1}^T w_k^\top \mathbf{A}_k w_k} \ . \tag{I.1}$$

Above, ① uses $\mathbf{Tr}(\mathbf{A}_T\Phi_{T-1}^{uu^\top}\mathbf{A}_T) = \mathbf{Tr}(\mathbf{A}_T^2\Phi_{T-1}^{uu^\top}) \leq \mathbf{Tr}(\mathbf{A}_T\Phi_{T-1}^{uu^\top})$ as well as $\Phi_{T-1}^{uu^\top} = \mathbf{Tr}(\Phi_{T-1}^{uu^\top})w_T w_T^\top$, and ② uses $1 + x \leq e^x$.

$$\mathbb{E}[\mathbf{Tr}(\Psi_1)] = \mathbb{E}\big[\mathbf{Tr}\big(\nu\nu^\top(\mathbf{I} + \eta\mathbf{A}_T)\Phi_{T-1}^{\mathbf{I}}(\mathbf{I} + \eta\mathbf{A}_T)\big)\big]$$
$$= \mathbb{E}[\mathbf{Tr}(\nu\nu^\top(\mathbf{I} + 2\eta\mathbf{A}_T)\Phi_{T-1}^{\mathbf{I}}) + \eta^2\nu^\top\mathbf{A}_T\Phi_{T-1}\mathbf{A}_T\nu]$$
$$\geq \mathbb{E}[\mathbf{Tr}(\nu\nu^\top(\mathbf{I} + 2\eta\mathbf{B})\Phi_{T-1}^{\mathbf{I}}) = (1 + 2\eta\lambda)\mathbb{E}[\nu^\top\Phi_{T-1}^{\mathbf{I}}\nu] \overset{\text{①}}{\geq} e^{2\eta\lambda - 2\eta^2\lambda^2}\mathbb{E}[\nu^\top\Phi_{T-1}^{\mathbf{I}}\nu]$$
$$\geq \cdots \geq e^{(2\eta\lambda - 2\eta^2\lambda^2)T} \cdot (\nu^\top\Phi_0^{\mathbf{I}}\nu) = e^{(2\eta\lambda - 2\eta^2\lambda^2)T} \ . \tag{I.2}$$

Above, ① uses $1 + 2x \geq e^{2x-2x^2}$ for $x \in [0,1]$, and the expectation is over $\mathbf{A}_1, \ldots, \mathbf{A}_T$.

$$\mathbb{E}\left[\mathbf{Tr}(\Psi_1^2)\right] = \mathbb{E}\left[\mathbf{Tr}((\mathbf{I}+\eta\mathbf{A}_1)^2\Psi_2(\mathbf{I}+\eta\mathbf{A}_1)^2\Psi_2)\right] \overset{①}{\leq} \mathbb{E}\left[\mathbf{Tr}((\mathbf{I}+\eta\mathbf{A}_1)^4\Psi_2^2)\right]$$
$$\overset{②}{\leq} \mathbb{E}\left[\mathbf{Tr}\big((\mathbf{I}+(4\eta+11\eta^2)\mathbf{A}_1)\Psi_2^2\big)\right] = \mathbf{Tr}\big((\mathbf{I}+(4\eta+11\eta^2)\mathbf{B})\,\mathbb{E}[\Psi_2^2]\big)$$
$$\leq e^{4\eta\lambda+11\eta^2\lambda}\,\mathbb{E}[\mathbf{Tr}(\Psi_2^2)] \leq \cdots \leq 3e^{(4\eta\lambda+11\eta^2\lambda)T}\ . \tag{I.3}$$

Above, ① uses the Lieb-Thirring inequality $\mathbf{Tr}(\mathbf{ABAB}) \leq \mathbf{Tr}(\mathbf{A}^2\mathbf{B}^2)$,[23] ② uses $(\mathbf{I}+\eta\mathbf{A}_1)^4 \preceq \mathbf{I}+(4\eta+11\eta^2)\mathbf{A}_1$, and the expectation is over $\mathbf{A}_1, \ldots, \mathbf{A}_T$.

Now, we can combine (I.2) and (I.3) and apply Chebyshev's inequality: for every $p \in (0,1)$

$$\Pr_{\mathbf{A}_1,\ldots,\mathbf{A}_T}\left[\mathbf{Tr}(\Psi_1) \leq e^{(2\eta\lambda-2\eta^2\lambda^2)T} - \frac{1}{\sqrt{p/2}}\sqrt{e^{(4\eta\lambda+11\eta^2\lambda)T} - (e^{(2\eta\lambda-2\eta^2\lambda^2)T})^2}\right] \leq p/2\ .$$

In other words, as long as $\lambda\eta^2 T \leq p/75$, we have with probability $\geq 1 - p/2$ over $\mathbf{A}_1, \ldots, \mathbf{A}_T$,

$$\mathbf{Tr}(\Psi_1) \geq e^{(2\eta\lambda-2\eta^2\lambda^2)T} \cdot (1 - (p/2)^{-1/2}\sqrt{e^{15\eta^2\lambda T} - 1}) \geq \frac{2}{3}e^{(2\eta\lambda-2\eta^2\lambda^2)T}\ . \tag{I.4}$$

Finally, denoting by $g \overset{\mathrm{def}}{=} (\mathbf{I}+\eta\mathbf{A}_1)\cdots(\mathbf{I}+\eta\mathbf{A}_T)\nu$, using tail bounds for chi-squared distribution [16], we have with probability at least $1 - p/2$ over the randomness of $u$:[24]

$$\|u\|_2^2 \leq d + O(\sqrt{d\log(1/p)}) \leq O(d + \log(1/p)) \quad \text{and} \quad (g^\top u)^2 \geq \Omega(p^{-2}\|g\|^2)\ .$$

Note the the later implies $\mathbf{Tr}(\Phi_T^{uu^\top}) \geq \nu^\top \Phi_T^{uu^\top}\nu = \|g^\top u\|^2 \geq \Omega(p^{-2}\|g\|^2) = \Omega(p^{-2})\mathbf{Tr}(\Psi_1)$. Plugging them into (I.1) and (I.4), we have with probability at least $1 - p$:

$$(2\eta+\eta^2)\sum_{k=1}^{T} w_k^\top \mathbf{A}_k w_k \geq 2\eta T\lambda - 2\eta^2\lambda^2 T - O\big(\log(d+\log(1/p))\big) - O(\log(1/p))\ ,$$

which after dividing both sides by $2\eta + \eta^2$ finishes the proof of Theorem 4. □

## J  A Simple Lower Bound for the $\lambda$-Refined Language

We sketch the proof that for the stochastic online eigenvector problem, for every $\lambda \in (0,1)$, there exists a constant $C > 0$, a PSD matrix $\mathbf{B}$ satisfying $\mathbf{B} \preceq \lambda\mathbf{I}$, and a distribution $\mathcal{D}$ of (even rank-1) matrices with spectral norm at most 1 and expectation equal to $\mathbf{B}$, such that for every learning algorithm Learner, the total regret must be at least $C \cdot \sqrt{\lambda T}$.

Such a lower bound naturally translates to the harder adversarial or oblivious settings. We prove this lower bound by reducing the problem to an information-theoretic lower bound that has appeared in our separate paper [8].

The lower bound in [8] states that, for every $1 \geq \lambda \geq \lambda_2 \geq 0$, there exists a PSD matrix $\mathbf{B}$ with the largest two eigenvalues being $\lambda$ and $\lambda_2$, and a distribution $\mathcal{D}$ of rank-1 matrices with spectral norm at most 1 and expectation equal to $\mathbf{D}$. Furthermore, for any algorithm Alg that takes $T$ samples from $\mathcal{D}$ and outputs a unit vector $v \in \mathbb{R}^d$, it must satisfy

$$\mathbb{E}[1 - \langle v, \nu_1\rangle^2] \geq \Omega\Big(\frac{\lambda}{(\lambda-\lambda_2)^2 T}\Big) \quad \text{for every } T \geq \Omega(\lambda/(\lambda-\lambda_2)^2)\ ,$$

---

[23] In fact, we do not need the full power of Lieb-Thirring here because one of the two matrices is rank-1.

[24] Chi-square distribution upper tail bound gives $\Pr[\|u\|_2^2 \geq (1+\alpha) \cdot d] \leq ((1+\alpha) \cdot e^{-\alpha})^{d/2}$ for $\alpha > 0$; choosing $\alpha = \Theta(d^{-1/2}\log^{1/2}(1/p))$ makes this probability $p$. The lower tail bound gives $\Pr[(g^\top u)^2 < \alpha \cdot \|g\|^2] \leq (\alpha \cdot e^{1-\alpha})^{1/2} \leq 2\sqrt{\alpha}$ for $\alpha \in [0,1]$.

where $\nu_1$ is the first eigenvector of $\mathbf{B}$, and the expectation is over the randomness of Alg and the $T$ samples from $\mathcal{D}$. After rewriting, we have

$$\mathbb{E}[v^\top \mathbf{B} v] \leq \mathbb{E}[\lambda \langle v, \nu_1 \rangle^2 + \lambda_2 (1 - \langle v, \nu_1 \rangle^2)] = \mathbb{E}[\lambda - (\lambda - \lambda_2)(1 - \langle v, \nu_1 \rangle^2)] \leq \lambda - \Omega\Big(\frac{\lambda}{(\lambda - \lambda_2)T}\Big) \ .$$

If we choose $\lambda_2$ such that $T = \Theta(\lambda/(\lambda - \lambda_2)^2)$, then the above inequality becomes

$$\mathbb{E}[v^\top \mathbf{B} v] \leq \lambda - \Omega(\sqrt{\lambda/T}) \ .$$

Finally, for any algorithm Learner for the stochastic online eigenvector problem, suppose Learner takes $T$ samples $\mathbf{A}_1, \ldots, \mathbf{A}_T$ from $\mathcal{D}$ and outputs unit vectors $v_1, \ldots, v_T$, we can define a corresponding algorithm Alg that outputs $v = v_k$ each with probability $1/T$. In this way, we have

$$\mathbb{E}\Big[\sum_{k=1}^T v_k^\top \mathbf{A}_k v_k\Big] = \mathbb{E}\Big[\sum_{k=1}^T v_k^\top \mathbf{B} v_k\Big] = T\,\mathbb{E}\big[v \mathbf{B} v\big] \leq \lambda T - \Omega(\sqrt{\lambda T}) \ .$$

In other words, the total regret of Learner must be at least $\Omega(\sqrt{\lambda T})$.